\documentclass[10pt,journal,compsoc]{IEEEtran}
\usepackage[nocompress]{cite}
\usepackage[pdftex]{graphicx}
\graphicspath{{figures/}{photos/}}
\DeclareGraphicsExtensions{.pdf,.jpeg,.png}
\usepackage{amsmath,bm}
\usepackage{algorithmic}
\usepackage{array}
\usepackage{url}
\usepackage[x11names]{xcolor}
\usepackage{tabulary,xfrac,enumitem}
\usepackage{booktabs}       %
\usepackage{amsfonts}       %
\usepackage{nicefrac}       %
\usepackage{microtype}      %
\usepackage{amssymb}
\usepackage{amsmath}
\usepackage{graphicx, caption, subcaption}
\usepackage{multirow}
\usepackage{textcomp}
\usepackage{makecell}
\usepackage{colortbl}
\usepackage{pifont}%
\usepackage{xspace}
\usepackage[ruled,noresetcount]{algorithm2e}
\usepackage[pagebackref=true,breaklinks=true,colorlinks,citecolor=citecolor,bookmarks=false]{hyperref}
\usepackage{diagbox}
\usepackage[capitalize]{cleveref}
\crefname{section}{Sec.}{Secs.}
\Crefname{section}{Section}{Sections}
\Crefname{table}{Table}{Tables}
\crefname{table}{Tab.}{Tabs.}

\usepackage{amsmath,amsfonts,bm}

\def\Eqref#1{Equation~\ref{#1}}
\def\eqref#1{\Eqref{#1}}

\def\1{\bm{1}}

\def\vs{{\bm{s}}}

\DeclareMathAlphabet{\mathsfit}{\encodingdefault}{\sfdefault}{m}{sl}
\SetMathAlphabet{\mathsfit}{bold}{\encodingdefault}{\sfdefault}{bx}{n}

\usepackage{subcaption}
\usepackage{bbm}
\usepackage{mathtools}
\usepackage{amsthm, amssymb}
\newtheorem{theorem}{Theorem}[section]

\def\1{\bm{1}}

\newcommand{\Id}{\bf Id}

\newcommand{\mc}{\mathcal}

\newcommand{\finntk}{\hat\Theta}
\newcommand{\infntk}{\Theta}

\definecolor{seagreen}{RGB}{62,187,163}
\definecolor{ZhuangGreen}{HTML}{d4fcd8}

\newcommand{\na}{\textcolor{gray}{-}}
\newcommand{\cmark}{\ding{51}\xspace}
\newcommand{\xmark}{\ding{55}\xspace}
\def \pzo {\phantom{0}}

\makeatletter
\def\@fnsymbol#1{\ensuremath{\ifcase#1\or \dagger\or \ddagger\or
   \mathsection\or \mathparagraph\or \|\or **\or \dagger\dagger
   \or \ddagger\ddagger \else\@ctrerr\fi}}
\makeatother

\usepackage{amsmath}
\usepackage{amssymb}
\usepackage{mathtools}
\usepackage{amsthm}
\usepackage{colortbl}

\definecolor{F7E0D5}{RGB}{240,235,255}
\colorlet{Light}{white!20!F7E0D5}
\colorlet{LightG}{white!70!lightgray}
\definecolor{citecolor}{RGB}{34,139,34}

\definecolor{newcolor}{rgb}{0.0, 0.0, 0.0}%

\newcommand{\new}[1]{{\color{newcolor}{#1}}}

\def \pzo {\phantom{0}} %

\makeatletter
\DeclareRobustCommand\onedot{\futurelet\@let@token\@onedot}
\def\@onedot{\ifx\@let@token.\else.\null\fi\xspace}

\def\eg{\emph{e.g}\onedot} 
\def\ie{\emph{i.e}\onedot} 
 
\def\etc{\emph{etc}\onedot} \def\vs{\emph{vs}\onedot}

\makeatother

\makeatletter
\newcommand{\thickhline}{%
    \noalign {\ifnum 0=`}\fi \hrule height 0.5pt
    \futurelet \reserved@a \@xhline
}
\makeatother

\theoremstyle{definition}
\newtheorem{definition}[theorem]{Definition}

\begin{document}
\title{Efficient Training of Large Vision Models via Advanced Automated Progressive Learning}

\author{Changlin~Li,
        Jiawei~Zhang,
        Sihao~Lin,
        Zongxin~Yang,
        Junwei~Liang, 
        Xiaodan~Liang,
        Xiaojun~Chang~\IEEEmembership{Senior~Member,~IEEE}%
\IEEEcompsocitemizethanks{%
\IEEEcompsocthanksitem C. Li, and X. Chang are with Australian Artificial Intelligence Institute, University of Technology Sydney. Email: changlinli.ai@gmail.com; xiaojun.chang@uts.edu.au.
\IEEEcompsocthanksitem J. Zhang is with North China Electric Power University. Email: zjw1637@gmail.com.
\IEEEcompsocthanksitem S. Lin is with School of Computing Technologies, RMIT
University. Email: linsihao6@gmail.com.
\IEEEcompsocthanksitem Z. Yang is with Harvard University.\\ Email:~zongxin\_yang@hms.harvard.edu.
\IEEEcompsocthanksitem J. Liang is with The Hong Kong University of Science and Technology (Guangzhou). He is also affiliated with HKUST CSE.\\Email: junweiliang1114@gmail.com.
\IEEEcompsocthanksitem X. Liang is with Sun Yat-sen University. Email: xdliang328@gmail.com.
\IEEEcompsocthanksitem Work partially done when X. Chang was a visiting professor at HKUST (Guangzhou).
}
\thanks{Corresponding author: Xiaojun Chang.}%
% \thanks{Code: \url{https://github.com/changlin31/AutoProg-Zero}}
}

\markboth{}%
{}

\IEEEtitleabstractindextext{%
\new{
\begin{abstract}
The rapid advancements in Large Vision Models (LVMs), such as Vision Transformers (ViTs) and diffusion models, have led to an increasing demand for computational resources, resulting in substantial financial and environmental costs. This growing challenge highlights the necessity of developing efficient training methods for LVMs. Progressive learning, a training strategy in which model capacity gradually increases during training, has shown potential in addressing these challenges. In this paper, we present an advanced automated progressive learning (AutoProg) framework for efficient training of LVMs. We begin by focusing on the pre-training of LVMs, using ViTs as a case study, and propose AutoProg-One, an AutoProg scheme featuring momentum growth (MoGrow) and a one-shot growth schedule search. Beyond pre-training, we extend our approach to tackle transfer learning and fine-tuning of LVMs. We expand the scope of AutoProg to cover a wider range of LVMs, including diffusion models. First, we introduce AutoProg-Zero, by enhancing the AutoProg framework with a novel zero-shot unfreezing schedule search, eliminating the need for one-shot supernet training. Second, we introduce a novel Unique Stage Identifier (SID) scheme to bridge the gap during network growth. These innovations, integrated with the core principles of AutoProg, offer a comprehensive solution for efficient training across various LVM scenarios. Extensive experiments show that AutoProg accelerates ViT pre-training by up to 1.85× on ImageNet and accelerates fine-tuning of diffusion models by up to 2.86×, with comparable or even higher performance. This work provides a robust and scalable approach to efficient training of LVMs, with potential applications in a wide range of vision tasks. Code: \url{https://github.com/changlin31/AutoProg-Zero}
\end{abstract}

}

}

\maketitle

\IEEEraisesectionheading{\section{Introduction}\label{sec:introduction}}

\new{
\IEEEPARstart{R}{ecent}  developments of Large Vision Models (LVMs) demonstrate the importance of model scale, dataset scale, and training scale. Two streams of models represent the development of LVMs, the representative discriminative models, Vision Transformers (ViTs), and the representative generative model, diffusion models.
With powerful high model capacity and large amounts of data, ViTs have dramatically improved the performance on many tasks in computer vision (CV) \cite{touvron2020deit, liu2021swin}.
The pioneering ViT model~\cite{dosovitskiy2021image}, scales the model size to 1,021 billion FLOPs, 250$\times$ larger than ResNet-50~\cite{he2016deep}.
Through pre-training on the large-scale JFT-3B dataset \cite{zhai2021scaling}, the ViT model, CoAtNet~\cite{dai2021coatnet}, reached remarkable performance, with about 8$\times$ training cost of the original ViT. %
For generative models, the recently popular Diffusion Transformer (DiT)~\cite{peebles2023scalable} achieves superior performance on the ImageNet class-conditional generation task. Its training requires 950 V100 GPU days on 256×256 images, and 1733 V100 GPU days on 512×512 images, as estimated by~\cite{xie2023difffit}.
The rapid growth in the training scale of LVMs inevitably leads to higher environmental costs. As shown in~\cref{tab:cost_of_vision_models}, recent breakthroughs of ViTs have come with a
considerable growth of carbon emissions. Therefore, it is crucial to make LVM training sustainable in terms of computational and energy consumption.

\begin{figure}
    \centering
    \includegraphics[width=.9\linewidth]{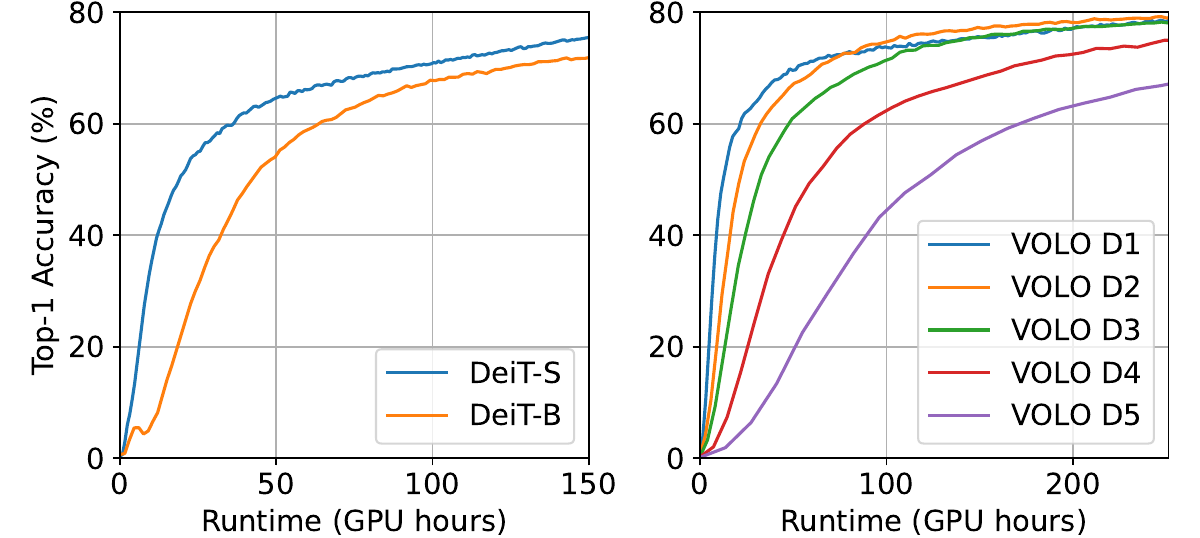}
    \caption{Accuracy of ViTs (DeiT~\cite{touvron2020deit}, VOLO~\cite{Yuan2021VOLOVO}) during training.
    Smaller ViTs converge faster in terms of runtime\protect\footnotemark.
    Models in the legend are sorted in ascending order of model size.}
    \label{fig:motivation1}
\end{figure}
\footnotetext{We refer runtime to the total GPU hours used in forward and backward pass of the model during training.}

\begin{figure*}[t]
    \centering
    \includegraphics[width=0.95\linewidth]{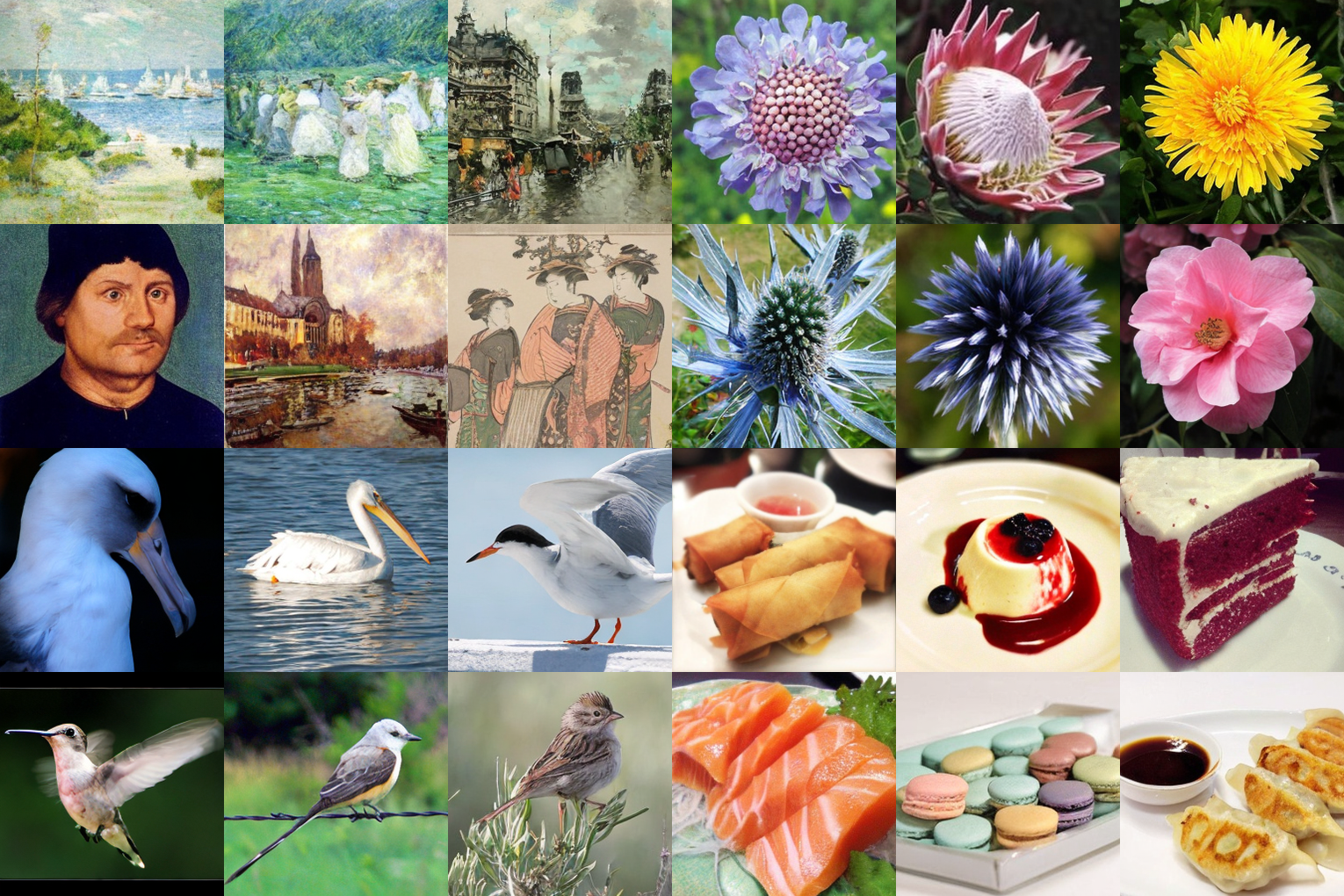}
    \caption{\new{
    \textbf{Generated images on multiple datasets with AutoProg-\textit{Zero} fine-tuning, which only takes \textit{0.39×} training time of full fine-tuning.} Resolution is 256×256.
    }}
    \label{fig:post}
\end{figure*}

In mainstream deep learning training schemes, all the network parameters participate in every training iteration. However, we empirically found that training only a small part of the parameters yields comparable performance in early training stages of ViTs. As shown in \cref{fig:motivation1},
smaller ViTs converge much faster in terms of runtime (though they would be eventually surpassed given enough training time). The above observation motivates us to rethink the efficiency bottlenecks of LVM training:
\textit{does every parameter, every input element need to participate in all the training steps?}

\begin{table}[t]
    \centering
    \scalebox{0.8}{
    \begin{tabular}{l|c|c}
        \toprule
         Model & CO$_2$e (lbs)\protect\footnotemark & ImageNet Acc. (\%) \\
         \midrule
         ResNet-50~\cite{he2016deep,dosovitskiy2021image} & 267 & 77.54\\
         \textcolor{gray}{BERT$_{base}$}~\cite{devlin2019bert} & \textcolor{gray}{1,438} & \textcolor{gray}{-} \\
         \textcolor{gray}{Avg person per year}~\cite{strubell2019energy} & \textcolor{gray}{11,023}& \textcolor{gray}{-} \\
         ViT-H/14~\cite{dosovitskiy2021image} & 22,793 & 88.55\\
         CoAtNet~\cite{dai2021coatnet} & 183,256 & 90.88\\
         \bottomrule
    \end{tabular}}
    \caption{The growth in training scale of vision models results in considerable growth of environmental costs. The CO$_2$e of human life and a language model, BERT~\cite{devlin2019bert} are also included for comparison. The results of ResNet-50, ViT-H/14 are from~\cite{dosovitskiy2021image}, and trained on JFT-300M~\cite{Sun2017RevisitingUE}. CoAtNet is trained on JFT-3B~\cite{zhai2021scaling}.} %
    \label{tab:cost_of_vision_models}
\end{table}
\footnotetext{CO$_2$ equivalent emissions (CO$_2$e) are calculated following~\cite{Patterson2021CarbonEA}, using U.S. average energy mix, \ie, 0.429 kg of CO$_2$e/KWh.}

The \textit{lottery ticket hypothesis}~\cite{frankle2019lottery} in the field of network pruning believes that a randomly-initialized, dense neural network contains a sub-network that can reach the performance of the full network after training for at most the same number of iterations.
Here, we make the \textit{Growing Ticket Hypothesis} of LVMs:
the performance of a Large Vision Model, can be reached by first training its sub-network, then the full network after properly growing (or unfreezing), with the same total training iterations. The proper growing (or unfreezing) schedule is the \textit{Growing Ticket} we need to find.
This hypothesis generalizes the \textit{lottery ticket hypothesis}~\cite{frankle2019lottery} by adding a finetuning procedure at the full model size,
changing its scenario from \textit{efficient inference} to \textit{efficient training}. By iteratively applying this hypothesis to the sub-network, we have the progressive learning scheme.

Recently, progressive learning has started showing its capability in accelerating model training.
In the field of NLP, progressive learning can reduce half of BERT pre-training time~\cite{Gong2019EfficientTO}.
Progressive learning also shows the ability to reduce the training cost for convolutional neural networks (CNNs)~\cite{Tan2021EfficientNetV2SM}.
However, these algorithms differ substantially from each other, and their generalization ability among architectures is not %
well studied.
For instance, we empirically observed that progressive stacking~\cite{Gong2019EfficientTO} could result in significant performance drop (about 1\%) on ViTs.

To this end, we take a practical step towards sustainable deep learning by generalizing and automating progressive learning on LVMs, including ViTs and diffusion models. To cover both the pre-training and fine-tuning of LVMs, we study them separately by using ViTs and diffusion models as the case study, respectively.

We focus on the efficient pre-training of ViTs as a representative case of LVMs. To begin, we establish a robust manual baseline for progressive learning in ViTs by developing a \textit{growth operator}. To evaluate the optimization process of this growth operator, we introduce a \textit{uniform linear growth schedule} that operates along two critical dimensions of ViTs: the number of patches and network depth. To address the challenges posed by model expansion during training, we propose a novel \textit{momentum growth} (\textit{MoGrow}) operator, which incorporates an effective momentum update scheme to smooth the transition as the model grows.
Furthermore, we introduce an innovative \textit{automated progressive learning} (\textit{AutoProg}) algorithm designed to accelerate training without compromising performance. AutoProg achieves this by dynamically adjusting the training workload in response to the model's growth. Specifically, we simplify the optimization of the growth schedule by framing it as a sub-network architecture optimization problem. To streamline this process, we propose a one-shot estimation method for evaluating sub-network performance, which leverages an \textit{elastic supernet}. We term this AutoProg algorithm with one-shot schedule search as \textit{AutoProg-One}. By recycling the parameters of the supernet, we significantly reduce the computational overhead associated with the search process.

Additionally, we expand the capabilities of AutoProg beyond just the pre-training of LVMs to also address the fine-tuning phase.
Fine-tuning is an essential step in the deployment of LVMs, adapting these general models to specific applications with minimal additional training. Recognizing the importance of this phase, we adapt the principles of progressive learning to make fine-tuning more efficient, ensuring that large models can be customized with lower computational costs. We also extend the scope of AutoProg to encompass a wider range of LVMs, including generative models. Therefore, we adopt diffusion models as a case study for efficient fine-tuning. To achieve this, we introduce several key innovations within the AutoProg framework. 
First, we introduced a progressive unfreezing scheme for efficient fine-tuning, which corresponds to the progressive growing scheme previously introduced for efficient pre-training. Progressive unfreezing reduces training overhead by freezing part of the model parameters without altering the architecture of the pre-trained model.
Second, we introduce a \textit{Unique Stage Identifier} (\textit{SID}) scheme designed to bridge the optimization gap during progressive unfreezing. This scheme minimizes the
fluctuations of the original ``dictionary'' of the diffusion model when switching training stages by adding a new ``vocabulary''
into the diffusion model’s ``dictionary''.
Third, we develop a novel zero-shot automated progressive learning method (\textit{AutoProg-\textit{Zero}}), which eliminates the need for one-shot supernet training and Neural Architecture Search (NAS) during AutoProg. More specifically, we introduce two different zero-shot metrics to evaluate the candidate learnable sub-network performance without training.

Our Advanced AutoProg framework has shown remarkable success in enhancing training efficiency by automatically determining  \textit{whether}, \textit{where} and \textit{how much} should the model grow or unfreeze during training. Through extensive experiments conducted across multiple datasets and a variety of LVM architectures, we demonstrated that this advanced automated progressive learning framework not only accelerates training but also preserves, and in some cases, even improves model performance.
In the case of ViTs, AutoProg-\textit{One} achieves a substantial acceleration of up to 1.85× during pre-training on the ImageNet dataset, while maintaining performance parity with traditional training methods. Moreover, when applied to the fine-tuning of diffusion models, AutoProg-\textit{Zero} delivers even more impressive results. It accelerates the transfer fine-tuning process of Stable Diffusion~\cite{rombach2022highresolution} and DiT by up to 2.86× and 2.56×, respectively, achieving comparable or superior performance relative to conventional approaches. \cref{fig:post} showcases the photo-realistic images generated using the efficient AutoProg training, achieving this with only 0.39× of the original training cost.
These results highlight the robustness and scalability of the AutoProg framework, making it a powerful tool for efficient training across a wide range of vision tasks. 

The ability to significantly reduce training time without sacrificing accuracy or performance positions AutoProg as a critical advancement in the field of LVMs, with broad potential applications in both research and industry.
Overall, our contributions are as follows:
\begin{itemize}
    \item 
    We establish a strong manual baseline for the progressive pre-training of ViTs by customizing a \textbf{progressive growing} space tailored to ViTs and introducing \textbf{MoGrow}, a momentum growth strategy designed to address the gap caused by model growth.
    \item
    We propose \textbf{AutoProg-\textit{One}}, an efficient training scheme aimed at achieving lossless acceleration by adaptively adjusting the growth schedule in real-time through one-shot search of candidate schedules.
    \item 
    Furthermore, we extend progressive learning to efficient fine-tuning, presenting \textbf{progressive unfreezing} for diffusion models. In addition, we introduce \textbf{SID}, a stage-wise prompt strategy to handle the transition challenges between training stages.
    \item
    We further extend AutoProg to efficient fine-tuning with \textbf{AutoProg-\textit{Zero}}, which enables lossless acceleration by automatically optimizing the unfreezing schedule on-the-fly through zero-shot estimation.
    \item While maintaining performance parity with traditional training methods, our Advanced AutoProg framework achieves remarkable pre-training acceleration (up to \textbf{1.85×}) for ViTs, and even more impressive acceleration (up to \textbf{2.86×}) for fine-tuning of diffusion models.
\end{itemize}

}

\section{Related Work}\label{sec:related_work}
\noindent\textbf{Progressive Learning.}
Early works on progressive learning~\cite{Fahlman1989TheCL,Lengell1996TrainingML,Hinton2006AFL,Bengio2006GreedyLT,Simonyan2015VeryDC,Smith2016GradualDO,Karras2018ProgressiveGO,Wang2017DeepGL} mainly focus on circumventing the training difficulty of deep networks.
Recently, as training costs of modern deep models are becoming formidably expensive, progressive learning starts to reveal its ability in \textit{efficient training}. %
Net2Net~\cite{Chen2016Net2NetAL} and Network Morphism \cite{Wei2016NetworkM,Wei2021ModularizedMO} studied how to accelerate large model training by properly initializing from a smaller model. %
In the field of NLP, many recent works accelerate BERT pre-training by progressively stacking layers~\cite{Gong2019EfficientTO,Li2020ShallowtoDeepTF,Yang2020ProgressivelyS2}, dropping layers~\cite{zhang2020accelerating} or growing in multiple network dimensions~\cite{Gu2021OnTT}. Similar frameworks have also been proposed for efficient training of other models~\cite{You2020L2GCNLA,Wang2021StackRecET}.
As these algorithms remain hand-designed and could perform poorly when transferred to other networks, we propose to automate the design process of progressive learning schemes.

\noindent\textbf{Automated Machine Learning.}
Automated Machine Learning (AutoML) aims to automate the design of model structures and learning methods %
from many aspects, including Neural Architecture Search (NAS)~\cite{zoph2016neural,baker2016designing,Tan2018MnasNetPN,liu2018progressive}, Hyper-parameter Optimization (HPO)~\cite{Bergstra2011AlgorithmsFH,Bergstra2012RandomSF}, AutoAugment~\cite{Cubuk2019AutoAugmentLA,tang2022learning}, AutoLoss~\cite{Wu2018LearningTT,Xu2019AutoLossLD,li2020autosegloss}, \etc By relaxing the bi-level optimization problem in AutoML, there emerges many \textit{efficient AutoML algorithms} such as weight-sharing NAS~\cite{Liu2018DARTSDA,Cai2018ProxylessNASDN,brock2017smash,pham2018enas,guo2020single,li2019blockwisely,peng2021pi,wang2023dna}, differentiable AutoAug~\cite{Li2020DADADA}, \etc These methods share network parameters in a jointly optimized \textit{supernet} for different candidate architectures or learning methods,
then rate each of these candidates according to its parameters inherited from the supernet. 

Attempts have also been made on \textit{automating progressive learning}.
AutoGrow~\cite{Wen2020AutoGrowAL} proposes to use a \textit{manually-tuned} progressive learning scheme to search for the optimal network depth, which is essentially a NAS method.
LipGrow~\cite{Dong2020TowardsAR} could be the closest one related to our work, which adaptively
decide the proper time to double the depth of CNNs on small-scale datasets, based on Lipschitz constraints. Unfortunately, LipGrow can not generalize easily to ViTs, as self-attention is not Lipschitz continuous~\cite{Kim2021TheLC}. In contrast, by improving over our conference version~\cite{li2022automated}, we solve the automated progressive learning problem from a traditional AutoML perspective, and fully automate the growing schedule by adaptively deciding \textit{whether}, \textit{where} and \textit{how much} to grow. Besides, our study is conducted directly on large-scale ImageNet dataset, in accord with practical application of efficient training.

\noindent\textbf{Elastic Networks.}
Elastic Networks, or anytime neural networks, are supernets executable with their sub-networks in various sizes, permitting instant and adaptive accuracy-efficiency trade-offs at runtime. Earlier works on Elastic Networks can be divided into \textit{Networks with elastic depth} \cite{larsson2016fractalnet,Huang2018MultiScaleDN,hu2019learning}, and \textit{networks with elastic width}~\cite{Lee2018AnytimeNP,yu2019slimmable,Yu2019UniversallySN}.
The success of elastic networks is followed by their two main applications, \textit{one-shot single-stage NAS}~ \cite{yu2019autoslim,Cai2020Once_for_All,Yu2020BigNASSU,chen2021autoformer} and \textit{dynamic inference}~\cite{Huang2018MultiScaleDN,Li2019ImprovedTF,li2021dynamic,wang2021not,li2021ds,jiang2023dynamic}, where emerges numerous elastic networks on \textit{multiple dimensions} (\eg, kernel size of CNNs~\cite{Cai2020Once_for_All,Yu2020BigNASSU}, head numbers~\cite{chen2021autoformer,hou2020dynabert} and patch size~\cite{wang2021not} of Transformers). From a new perspective, we present an elastic Transformer serving as a sub-network performance estimator during growth for automated progressive learning.%

\new{
\noindent\textbf{Diffusion Models.}
The Denoising Diffusion Probabilistic Model (DDPM)\cite{ho2020denoising} highlighted the effectiveness of U-Net-based architectures.
Building on this, Score-based generative models\cite{song2021scorebased} introduced a novel framework that integrates denoising and score-matching techniques.
Diffusion Transformer (DiT)~\cite{peebles2023scalable} represents a significant innovation by employing a pure Transformer as the backbone network.
The advent of multimodal models such as CLIP~\cite{radford2021learning} and DALL·E~\cite{ramesh2021zero},
has substantially elevated the capabilities of Text-to-Image (T2I) generation. Large-scale diffusion models, including Imagen~\cite{saharia2022photorealistic}, DALL-E2~\cite{ramesh2022hierarchical}, and Stable Diffusion~\cite{rombach2022highresolution}, have set new standards in T2I generation.

\noindent\textbf{Efficient Fine-tuning.}
Efficient fine-tuning has emerged as a critical research area in the context of large models.
Key approaches in this domain include parameter-efficient fine-tuning methods like Partial Parameter Tuning~\cite{zaken2021bitfit,xie2023difffit}, Adapter Tuning~\cite{hu2021lora} and Prompt Tuning~\cite{li2021prefix}. For instance, BitFit~\cite{zaken2021bitfit} adjusts only the bias terms of each linear projection, and DiffFit~\cite{xie2023difffit} fine-tunes both the bias term and scaling factor. Low-Rank Adaptation (LoRA)~\cite{hu2021lora} 
employs adapters for fine-tuning, drastically reducing the number of trainable parameters.
Prefix tuning~\cite{li2021prefix} inserts trainable tokens before the input tokens at each layer of the self-attention module. However, parameter-efficient fine-tuning methods can sometimes underperform on diffusion models, as they do not optimize the whole network for the new task. In contrast, AutoProg efficiently optimizes all the network parameters, ensuring better performance.%

}

\begin{figure*}[t]
    \centering
    \includegraphics[width=.95\linewidth]{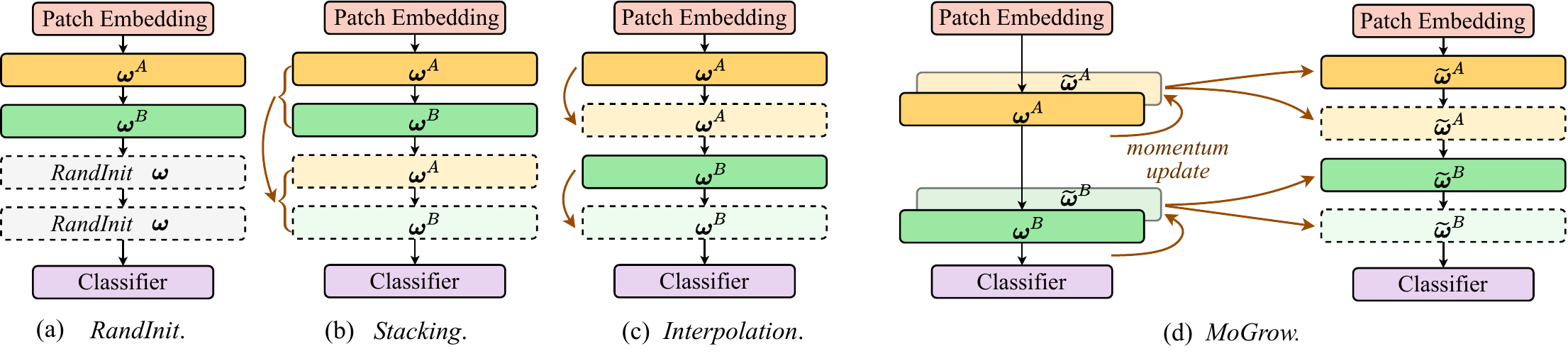}
    \caption{Variants of the growth operator $\bm\zeta$. $\bm\omega^\textit{A}$ and $\bm\omega^\textit{B}$ denote the parameters of two Transformer blocks in the original small network $\bm\psi_{k-1}$. (a) \textit{RandInit} randomly initializes newly added layers; (b) \textit{Stacking} duplicates the original layers and directly stacks the duplicated ones on top of them; (c) \textit{Interpolation} interpolate new layers of $\bm\psi_\ell$ in between original ones and copy the weights from their nearest neighbor in $\bm\psi_{k-1}$. (d) Our proposed \textit{MoGrow} is build upon \textit{Interpolation}, by coping parameters $\widetilde{\bm\omega}$ from the momentum updated ensemble of $\bm\psi_{k-1}$.}
    \label{fig:growth_operators}
\end{figure*}

\section{Automated Progressive Pre-training of Vision Transformers}\label{sec:3}

\subsection{Progressive Learning for Efficient Pre-training of Vision Transformers}\label{sec:3.1}

In this section, we aim to develop a strong manual baseline for progressive learning of ViTs. 
We start by formulating progressive learning with its two main factors, \textit{growth schedule} and \textit{growth operator} in Sec.~\ref{sec:Prog}. Then, we present the growth space that we use in Sec.~\ref{sec:GS}. Finally, we explore the most suitable growth operator of ViTs in Sec.~\ref{sec:GO}.

\noindent\textbf{Notations.} We denote scalars, tensors 
and sets (or sequences) using lowercase, bold lowercase and uppercase letters (\eg, $n$, $\bm x$ and $\Psi$). For simplicity, we use $\{\bm x_n\}$ to denote the set $\{\bm x_n\}^{|n|}_{n=1}$ with cardinality $|n|$, similarly for a sequence $\left(\bm x_n\right)^{|n|}_{n=1}$. Please refer to \cref{tab:notations}
for a vis-to-vis explanation of the notations we used.
\subsubsection{Progressive Learning} \label{sec:Prog}

Progressive learning gradually increases the training overload by growing among its sub-networks following
a \textit{growth schedule} $\Psi$, which can be denoted by a sequence of sub-networks with increasing sizes for all the training epochs~$t$.
In practice, to ensure the network is sufficiently optimized after each growth, it is a common practice~\cite{Yang2020ProgressivelyS2,Gu2021OnTT,Tan2021EfficientNetV2SM} to divide the whole training process into $|k|$ equispaced stages with $\tau = |t|/|k|$ epochs in each stage. Thus, the growth schedule can be denoted as ${\Psi = \Big(\bm{\psi}_k\Big)_{k=1}^{|k|}}$; the final one is always the complete model.
Note that stages with different lengths can be achieved by using the same $\bm\psi$ in different numbers of consecutive stages, \eg, ${\Psi = (\bm\psi_a, \bm\psi_b, \bm\psi_b)}$, where $\bm\psi_a, \bm\psi_b$ are two different sub-networks.

When growing a sub-network to a larger one, the parameters of the larger sub-network are initialized by a \textit{growth operator} $\bm\zeta$, which is a reparameterization function that maps the weights $\bm\omega_{k-1}$ of a smaller network in the $k-1$ stage to $\bm\omega_k$ of a larger one in stage $k$ by ${\bm\omega_k=\bm\zeta(\bm\omega_{k-1})}$. %
\begin{table}[t]
    \centering
    \footnotesize
    \setlength{\tabcolsep}{4pt}
    \begin{tabular}{l|c|l}
    \toprule
    Notation        & Type & Description  \\
    \midrule
    $s$, $|s|$      & scalar    & training step, total training steps \\    $t$, $|t|$      & scalar    & training epoch, total training epochs \\
    $k$, $|k|$      & scalar    & training stage, total training stages \\
    $\tau$          & scalar    & epochs (or steps) per stage \\
    $\Psi$          & sequence  & growth schedule \\
    $\bm\zeta$      & function   & growth operator\\
    $\bm\psi$       &   network        & sub-network \\
    $\Phi$          &  network         & supernet \\
    $\bm\omega$, $|\bm\omega|$     &  parameter         & network parameters, number of parameters\\
    $\Omega$, $\Lambda$         &   set    & growth space, partial growth space\\
    $*$, $\star$       &    notation       & optimal, relaxed (estimated) optimal\\
    $\mathcal{L}, \mathcal{T}, \mathcal{H}$ & function & loss, runtime, zero-shot metrics\\
    \bottomrule
    \end{tabular}
    \caption{Notations describing progressive learning and automated progressive learning.}
    \label{tab:notations}
\end{table}

Let $\mathcal{L}$ be the target loss function,
and $\mathcal{T}$ be the total runtime; then progressive learning can be formulated as:
\begin{equation}\label{eq:prog}
    \mathop{\min}_{\bm\omega, \Psi, \bm\zeta}\big\{\mathcal{L}(\bm\omega, \Psi, \bm\zeta), \mathcal{T}(\Psi)\big\},
\end{equation}
where $\bm\omega$ denotes the parameters of sampled sub-networks during the optimization.
Growth schedule $\Psi$ and growth operator $\bm\zeta$ have been explored for language Transformers~\cite{Gong2019EfficientTO,Gu2021OnTT}. However, ViTs differ considerably from their linguistic counterparts. The huge difference on task objective, data distribution and network architecture could lead to drastic difference in optimal $\Psi$ and $\bm\zeta$. In the following parts of this section, we mainly study the growth operator $\bm\zeta$ for ViTs by fixing $\Psi$ as a \textit{uniform linear schedule} in a \textit{growth space} $\Omega$, and leave automatic exploration of $\Psi$ to~\cref{sec:autoprogone}.

\subsubsection{Growth Space in Vision Transformers} \label{sec:GS}

The model capacity of ViTs are controlled by many factors, such as number of patches, network depth, embedding dimensions, MLP ratio,
\etc. In analogy to previous discoveries on fast compound model scaling~\cite{Dollr2021FastAA}, we empirically find that reducing network width (\eg, embedding dimensions) yields relatively smaller wall-time acceleration on modern GPUs when comparing at the same \textit{flops}. Thus, we mainly study \textit{number of patchs ($n^2$)} and \textit{network depth~($l$)}, leaving other dimensions for future works.

\noindent\textbf{Number of Patches.} %
Given patch size $p\times p$, input size ${r\times r}$, the number of patches $n\times n$ is determined by $n^2 = r^2/p^2$. Thus, by fixing the patch size,
reducing \textit{number of patches} can be simply achieved by %
down-sampling the input image. 
However, in ViTs, the size of positional encoding is related to~$n$. To overcome this limitation, we adaptively interpolate the positional encoding to match with~$n$.

\noindent\textbf{Network Depth.}
Network depth ($l$) is the number of Transformer blocks or its variants (\eg, Outlooker blocks~\cite{Yuan2021VOLOVO}). %

\noindent\textbf{Uniform Linear Growth Schedule.}
To ablate the optimization of growth operator $\bm\zeta$, we fix growth schedule $\Psi$ as a \textit{uniform linear growth schedule}.  %
To be specific, \textit{``uniform''} means that all the dimensions (\ie, $n$ and $l$) are scaled by the same ratio $r_t$ at the $t$-th epoch; \textit{``linear''} means that the ratio~$r$
grows linearly from $r_1$~to~$1$. %
This manual schedule has only one hyper-parameter, the initial scaling ratio~$s_1$, which is set to $0.5$ by default. 
With this fixed $\Psi$, the optimization of progressive learning in \cref{eq:prog} is simplified to:
\begin{equation}\label{eq:zeta}
    \mathop{\min}_{\bm\omega, \bm\zeta}\mathcal{L}(\bm\omega, \bm\zeta),
\end{equation}
which enables direct optimization of $\bm\zeta$ by comparing the final accuracy after training with different $\bm\zeta$.
\subsubsection{On the Growth of Vision Transformers} \label{sec:GO}

\cref{fig:growth_operators}
(a)-(c) depict the main variants of the growth operator $\bm\zeta$ that we compare, which cover choices from a wide range of the previous works, including \textit{RandInit}~\cite{Simonyan2015VeryDC}, \textit{Stacking}~\cite{Gong2019EfficientTO} and \textit{Interpolation}~\cite{chang2018multi,Dong2020TowardsAR}. %
More formal definitions of these schemes can be found in the supplementary material.
Our empirical comparison (in~\cref{sec:exp_ablation}) shows Interpolation growth is the most suitable scheme for ViTs. %

Unfortunately, growing by Interpolation changes the original function of the network.
In practice, function perturbation brought by growth can result in significant performance drop, which is hardly recovered in subsequent training steps. %
Early works advocate for function-preserving growth operators~\cite{Chen2016Net2NetAL,Wei2016NetworkM}, which we denote by \textit{Identity}. However, we empirically found growing by Identity greatly harms the performance on ViTs (see~\cref{sec:exp_ablation}).
Differently, we propose a growth operator, named \textit{Momentum Growth (MoGrow)}, to bridge the gap brought by model growth.

\begin{figure*}[t]
    \centering
    \includegraphics[width=.95\linewidth]{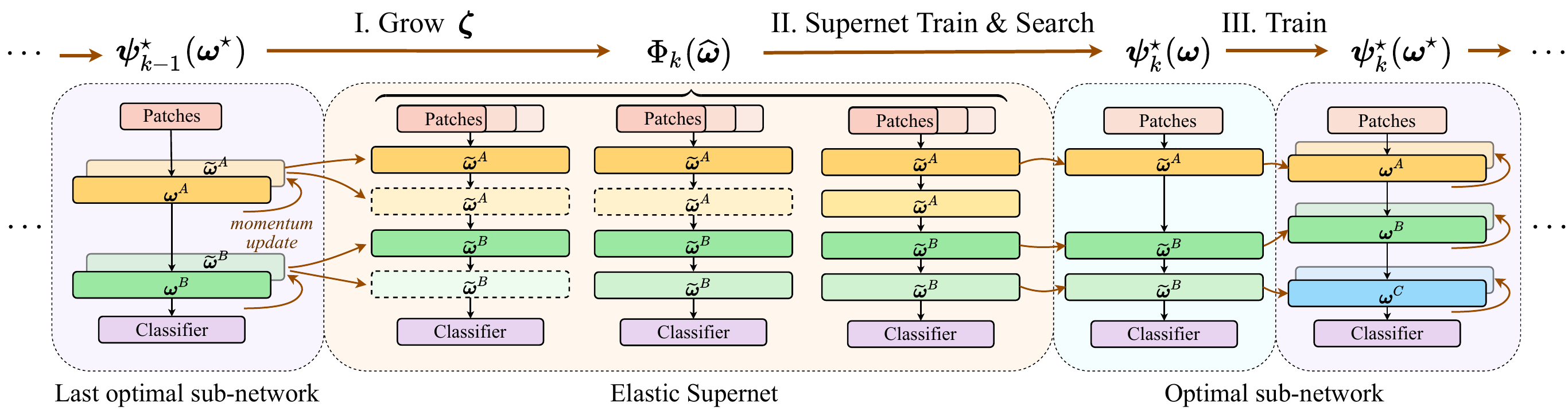}
    \caption{Pipeline of the $k$-th stage of AutoProg-\textit{One}. In the beginning of the stage, the last optimal sub-network $\bm\psi^\star_{k-1}$ first grows to the Elastic Supernet $\Phi_k$ by $\widehat{\bm\omega} = \bm\zeta(\bm\omega^\star)$; then, we search for the optimal sub-network $\bm\psi^\star_{k}$ after supernet training; finally, the sub-network is trained in the remaining epochs of this stage. The whole process of AutoProg-\textit{One} is summarized in \cref{alg:autoprog}.}
    \label{fig:autoprog}
\end{figure*}

\noindent\textbf{Momentum Growth (MoGrow).}
In recent years, a growing number of self-supervised~\cite{BYOL,guo2020bootstrap,He2020MomentumCF} and semi-supervised~\cite{laine2016temporal,tarvainen2017mean} methods learn knowledge from the historical ensemble of the network. Inspired by this, we propose to transfer knowledge from a \textit{momentum network} during growth. During training of the last stage (stage ${k-1}$), this momentum network has the same architecture with $\bm\psi_{k-1}$ and its parameters $\bm{\widetilde{\omega}}_{k-1}$ are updated with the online parameters $\bm\omega_{k-1}$ in every training step by:
\begin{equation}
    \bm{\widetilde{\omega}}_{k-1} \leftarrow m\bm{\widetilde{\omega}}_{k-1} + (1-m)\bm\omega_{k-1},
\end{equation}
where $m$ is a momentum coefficient set to $0.998$. As the the momentum network usually has better generalization ability and better performance
during training, loading parameters from the momentum network would help the model bypass the function perturbation gap. As shown in \cref{fig:growth_operators} (d),
MoGrow is proposed upon Interpolation growth by maintaining a momentum network, and directly copying from it when performing network growth. MoGrow operator $\bm\zeta_\textit{MoGrow}$ can be simply defined as:
\begin{equation}
    \bm\omega_{k} = \bm\zeta_\textit{MoGrow}(\bm\omega_{k-1}) = \bm\zeta_\textit{Interpolation}\big(\bm{\widetilde{\omega}}_{k-1}\big).
\end{equation}

\subsection{Automated Progressive Learning for Pre-training}\label{sec:autoprogone}

\begin{algorithm}[t]
\footnotesize
\SetAlgoLined
\textbf{Input:}\\  
$\bm\zeta$:~the~growth~operator;\\
$|t|$:~total~training~epochs;
$\tau$:~epochs per stage;\\
Random initialize parameters $\bm\omega$;\\
\For {$t \in [1,|t|]$}{
    \If{~~$t=N\tau,~~N\in\mathbb{N_{+}}$~~}{
    Switch optimizers to \textit{Elastic Supernet}~~$\Phi$;\\
    Initialize supernet parameters~~$\widehat{\bm\omega}\leftarrow\bm\zeta(\bm\omega)$;\\}{}
    \If{~~$t=N\tau + 2,~~N\in\mathbb{N_{+}}$~~}{
    Search for the \textit{optimal sub-network}~$\bm\psi^\star$ by \cref{eq:objective_3};\\
    Switch to the \textit{optimal sub-network}~~$\bm\psi\leftarrow\bm\psi^\star$;\\
    Inherit parameters from the supernet~~$\bm\omega\leftarrow\widehat{\bm\omega}$;}{}
    Train $\bm\psi(\bm\omega)$ or supernet $\Phi(\widehat{\bm\omega})$ over all the training data.
}
\caption{AutoProg-\textit{One} for Pre-training}
\label{alg:autoprog}
\end{algorithm}

In this section, we focus on optimizing the growth schedule $\Psi$ by fixing the growth operator as $\bm\zeta_\textit{MoGrow}$. We first formulate the multi-objective optimization problem of $\Psi$, then propose our solution, called \textit{AutoProg}, which is introduced in detail by its two estimation steps in \cref{sec:subnet} and \cref{sec:supernet}.%
\subsubsection{Problem Formulation}
The problem of designing growth schedule $\Psi$ for efficient training is a multi-objective optimization problem~\cite{deb2014multi}. By fixing $\bm\zeta$ in \cref{eq:prog} as our proposed $\bm\zeta_\textit{MoGrow}$, the objective of designing growth schedule $\Psi$ is: 
\begin{equation}\label{eq:psi}
    \mathop{\min}_{\bm\omega, \Psi}\big\{\mathcal{L}(\bm\omega, \Psi), \mathcal{T}(\Psi)\big\}.
\end{equation}

Note that multi-objective optimization problem has a set of Pareto optimal~\cite{deb2014multi} solutions which can be approximated using customized weighted product, a common practice used in previous Auto-ML algorithms~\cite{Tan2018MnasNetPN,Tan2019EfficientNetRM}. In the scenario of progressive learning, the optimization objective can be defined as:
\begin{equation}\label{eq:psi_product}
    \mathop{\min}_{\bm\omega, \Psi}\mathcal{L}(\bm\omega, \Psi)\cdot \mathcal{T}(\Psi)^\alpha,
\end{equation}
where $\alpha$ is a balancing factor dynamically chosen by balancing the scores for all the candidate sub-networks. 
\subsubsection{Automated Progressive Learning by Optimizing Sub-Network Architectures}\label{sec:subnet}
Denoting $|\bm\psi|$ the number of candidate sub-networks, and $|k|$ the number of stages, the number of candidate growth schedule is thus $|\bm{\psi}|^{|k|}$. As optimization of \cref{eq:psi_product} contains optimization of network parameters $\bm\omega$, to get the final loss, a full $|t|$ epochs training with growth schedule $\Psi$ is required:
\begin{equation}
\label{eq:objective}
\begin{aligned}
    \Psi^* = \mathop{\arg\min}_{\Psi} \mathcal{L}\big(\bm{\omega}^*(\Psi); \bm{x}\big) \cdot \mathcal{T}(\Psi) ^{\alpha},\\
    {\text{s.t.}}~~~~\bm{\omega}^*(\Psi) = \mathop{\arg\min}_{\bm{\omega}} \mathcal{L}(\Psi, {\bm{\omega}}; \bm{x}).
\end{aligned}
\end{equation}
Thus, performing an extensive search over the higher level factor $\Psi$ in this bi-level optimization problem has complexity $\mathcal{O}(|\bm{\psi}|^{|k|}\cdot|t|)$. Its expensive cost %
deviates from the original intention of efficient training. 

To reduce the search cost, we relax the original objective of growth schedule search to progressively deciding \textit{whether}, \textit{where} and \textit{how much} should the model grow, by
searching the optimal sub-network architecture $\bm\psi^*_k$ in each stage $k$. Thus, the relaxed optimal growth schedule can be denoted as ${\Psi^\star = \Big(\bm{\psi}^*_k\Big)_{k=1}^{|k|}}$. %

We empirically found that the network parameters adapt quickly after growth and are already stable after one epoch of training. %
To make a good tradeoff between accuracy and training speed,
we estimate the performance of each sub-network $\bm{\psi}$ in each stage by their training loss after the first \textit{two} training epochs in this stage. 
Denoting $\bm{\omega}^\star$ the sub-network parameters obtained by two epochs of training, the optimal sub-network can be searched by:
\begin{equation}
\label{eq:objective_2}
\begin{aligned}
    &\bm\psi_k^* = \mathop{\arg\min}_{\bm{\psi}_k\in\Lambda_k}\mathcal{L}\big(\bm{\omega}^\star(\bm{\psi}_k);\bm{x}\big) \cdot \mathcal{T}(\bm{\psi}_k)^\alpha,\\
    \text{where}~~~~&\Lambda_k = \Big\{\bm\psi\in\Omega~\Big|~|\bm\omega(\bm\psi)| \geq |\bm\omega(\bm\psi^*_{k-1})|\Big\}.
\end{aligned}
\end{equation}
$\Lambda_k$ denotes the growth space of the $k$-th stage,
containing all the sub-networks that are larger than or equal to the previous optimal sub-network $\bm\psi^*_{k-1}$ in terms of the number of parameters $|\bm\omega|$.

Overall, by relaxing the original optimization problem in \cref{eq:objective} to \cref{eq:objective_2}, we only have to train each of the ${|\Lambda_k|}$ sub-networks for two epochs in each of the $|k|$ stages. Thus, the search complexity is reduced exponentially from $\mathcal{O}(|\bm{\psi}|^{|k|}\cdot|t|)$ to $\mathcal{O}(|\Lambda_k|\cdot|k|)$, where $|\Lambda_k|\leq|\bm{\psi}|$ and $|k|\leq|t|$.

\subsubsection{One-shot Estimation of Sub-Network Performance via Elastic Supernet}\label{sec:supernet}

Though we relax the optimization problem with significant search cost reduction,
obtaining $\bm\omega^\star$ in \cref{eq:objective_2} still takes ${2|\Lambda_k|}$ epochs for each stage, which will bring huge searching overhead to the progressive learning. %
The inefficiency of loss prediction is caused by the repeated training of sub-networks weight $\bm{\omega}$, with bi-level optimization being its nature. %
To circumvent this problem, 
we propose to share and jointly optimize sub-network parameters in $\Lambda_k$ via an \textit{Elastic Supernet with Interpolation}.

\begin{figure*}[t]
    \centering
    \includegraphics[width=.95\linewidth]{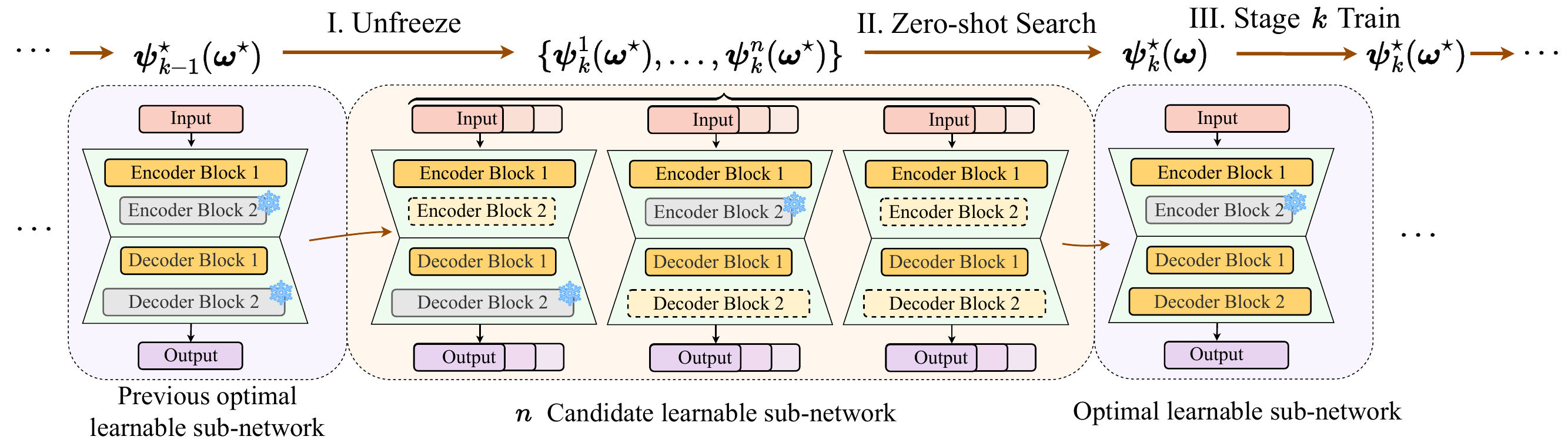}
    \caption{Pipeline of the $k$-th stage of AutoProg-\textit{Zero}. In the beginning of the stage, different candidate learnable sub-networks unfreeze; then, we search for the optimal learnable sub-network $\bm\psi^\star_{k}$ according to zero-shot metrics; finally, the sub-network unfreezes and is trained in this stage. The whole process of AutoProg-\textit{Zero} is summarized in \cref{alg:autoprogzero}.}
    \label{fig:autoprogzero}
\end{figure*}

\noindent\textbf{Elastic Supernet with Interpolation.}
An Elastic Supernet $\Phi({\widehat{\bm\omega}})$ is a \textit{weight-nesting} network parameterized by $\widehat{\bm\omega}$, and is able to execute with its sub-networks $\{\bm\psi\}$. Here, we give the formal definition of \textit{weight-nesting}:
\begin{definition}
(weight-nesting) For any pair of sub-networks $\bm\psi_a(\bm\omega_a)$ and $\bm\psi_b(\bm\omega_b)$ in supernet $\Phi$, where $|\bm\omega_a| \leq |\bm\omega_b|$, if $\bm\omega_a \subseteq \bm\omega_b$ is always satisfied, then $\Phi$ is \textbf{weight-nesting}.
\end{definition}
In previous works, a network with elastic depth is usually achieved by using the first layers to form sub-networks~\cite{Huang2018MultiScaleDN,Yu2020BigNASSU,chen2021autoformer}. However, using this scheme after growing by \textit{Interpolation} or \textit{MoGrow} will cause inconsistency between expected sub-networks after growth and sub-networks in $\Phi$.

To solve this issue, we present an \textit{Elastic Supernet with Interpolation}, with optionally activated layers interpolated in between always activated ones. 
As shown in \cref{fig:autoprog},
beginning from the smaller network in the last stage $\bm\psi^\star_{k-1}$, sub-networks in $\Phi$ are formed by inserting layers in between the original layers of $\bm\psi^\star_{k-1}$ (starting from the final layers), until reaching the largest sub-network in $\Lambda_k$. 

\noindent\textbf{Training and Searching via Elastic Supernet.} 
By nesting parameters of all the candidate sub-networks in the Elastic supernet $\Phi$, the optimization of $\bm\omega$ is disentangled from $\bm\psi$. Thus, \cref{eq:objective_2} is further relaxed to:
\begin{equation}\label{eq:objective_3}
\begin{aligned}
&\bm\psi_k^\star = \mathop{\arg\min}_{\bm\psi_k\in\Lambda_k} \mathcal{L}\big(\widehat{\bm\omega}^\star; \bm{x}\big) \cdot \mathcal{T}(\bm\psi_k) ^{\alpha},\\
{\text{s.t.}}~~~~&\widehat{\bm\omega}^{\star} = \mathop{\arg\min}^{~}_{\widehat{\bm\omega}}\mathbb{E}_{\bm\psi_k\in \Lambda_k}[\mathcal{L}(\bm\psi_k, \widehat{\bm\omega}; \bm{x})],
\end{aligned}
\end{equation}
where the optimal nested parameters $\widehat{\bm\omega}^\star$ can be obtained by one-shot training of $\Phi$ for two epochs. For efficiency, we train $\Phi$ by randomly sampling only one of its sub-networks in each step (following~\cite{chen2021autoformer}), instead of four in~\cite{Yu2019UniversallySN,yu2019autoslim,Yu2020BigNASSU}. 

After training all the candidate sub-networks in the Elastic Supernet $\Phi$ concurrently for two epochs, we have the adapted supernet parameters $\widehat{\bm\omega}^\star$ that can be used to estimate the real performance of the sub-networks (\ie performance when trained in isolation). As the sub-network grow space $\Lambda_k$ in each stage is relatively small,
we can directly perform traversal search in $\Lambda_k$, by testing its training loss with a small subset of the training data. We use fixed data augmentation to ensure fair comparison, following
\cite{Li2021BossNASEH}. Benefiting from parameter nesting and one-shot training of all the sub-networks in $\Lambda_k$, the search complexity is further reduced from $\mathcal{O}(|\Lambda_k|\cdot|k|)$ to $\mathcal{O}(|k|)$.

\noindent\textbf{Weight Recycling.}
Benefiting from synergy of different sub-networks, the supernet converges at a comparable~speed to training these sub-networks in isolation.
Similar phenomenon can be observed in network regularization~\cite{Srivastava2014DropoutAS,Huang2016DeepNW}, network augmentation~\cite{Cai2021NetworkAF}, and previous elastic models~\cite{yu2019slimmable,Yu2020BigNASSU,chen2021autoformer}. Motivated by this, the searched sub-network directly inherits its parameters in the supernet to continue training. %
Benefiting from this \textit{weight recycling} scheme, AutoProg-\textit{One} has \textit{no} extra searching epochs, since the supernet training epochs are parts of the whole training process. 
Moreover, as sampled sub-networks are faster than the full network, these supernet training epochs take less time than the original training epochs.
Thus, the searching cost is directly reduced from $\mathcal{O}(|k|)$ to \textbf{\textit{zero}}.

\new{

\section{Automated Progressive Fine-tuning of Diffusion Models}
In this section, we set to solve the problem of efficient fine-tuning via automated progressive learning. We use diffusion models as a case study. We start by developing a strong manual baseline for progressive fine-tuning. Then, we present AutoProg-\textit{Zero} for efficient fine-tuning.

\subsection{Progressive Learning for Efficient Fine-tuning of Diffusion Models}

\subsubsection{Progressive Fine-tuning}
Current computer vision tasks benefits a lot from adapting large pre-trained models through fine-tuning. Progressive learning introduced previously can only be applied on the pre-training phase of vision models. The efficiency issue of fine-tuning large pre-trained models remains unsolved. In this section, we set to solve this issue by generalizing progressive learning to efficient fine-tuning.

As introduced previously, progressive pre-training gradually increases the training overload by growing among its sub-networks. However, training by routing through a sub-network during fine-tuning can significantly harm the performance of a pre-trained LVM. To achieve progressive fine-tuning, we can prune or remove part of the network by ranking the importance of the pre-trained parameters and then reverse this process by progressive growing. However, such a process could still harm the performance of the pre-trained network and may increase the overall training overload due to the extra evaluation and ranking process of the pre-trained parameters. Instead of progressive growing, we seek a simpler yet more efficient way inspired by parameter-efficient fine-tuning schemes, namely \textit{progressive unfreezing}.

\noindent\textbf{Progressive Unfreezing.}
Progressive efficient fine-tuning gradually increases the training overload by first freezing all the parameters in the pre-trained models, then gradually unfreezing the parameters in the model following a \textit{unfreezing schedule} $\Psi$, in analogy to the growth schedule in progressive growing for efficient pretraining. Each status of the unfreezing schedule can be denoted as a network with different learnable settings, $e\in \{\textit{learnable}, \textit{frozen}\}$ for all the parameters. Similar to progressive pre-training, we divide the whole training process into $|k|$ equispaced stages.
The \textit{unfreezing schedule} $\Psi$ can then be denoted by a sequence of networks with increasing trainable parameters for all training stages. To align with
progressive pre-training, we denote networks with different learnable parameter settings in different 
stages using \textit{learnable sub-networks} $\bm{\psi}$.
Thus, the unfreezing schedule can be denoted as ${\Psi = \Big(\bm{\psi}_k\Big)_{k=1}^{|k|}}$; the final one is always the fully learnable model.

\subsubsection{Bridging the Gap in Progressive Fine-tuning with Unique Stage Identifier}
During progressive fine-tuning, when switching unfreezing stages, the input resolution and learnable sub-networks are changed substantially. Such a large distribution shift of input and optimization gap caused by unfreezing network parameters may lead to large fluctuations in the training dynamics during fine-tuning. Inspired by the subject-driven customization of diffusion models~\cite{ruiz2023dreambooth}, we propose the \textit{Unique Stage Identifier (SID)} to bridge the gap between different growing stages by customizing diffusion models in each unfreezing stage.

\noindent\textbf{Unique Stage Identifier (SID).}
Our goal is to minimize the fluctuations of the original ``dictionary'' of the diffusion model when switching training stages by adding a new (stage identifier [SID], training stage $k$) pair into the diffusion model's ``dictionary''. For text-to-image diffusion models, we label all input images corresponding to the training stage as ``a [class noun], [SID] stage'', where [SID] is a unique identifier associated with the training stage, and [class noun] is a class descriptor relevant to the input sample (\eg, flowers, birds, etc.). The class descriptor is usually given by the label or caption in the dataset.
Class-conditional diffusion models take a class embedding, instead of the text, as the condition. The class embedding for each class can be denoted as $\mathcal{C}\in\mathbb{R}^{1\times\textit{dim}}$, where $\textit{dim}$ is the hidden dimension of the class embeddings.
For these models, a learnable stage embedding $\textit{SID}\in\mathbb{R}^{1\times\textit{dim}}$ is added to the class embedding as the Unique Stage Identifier. For both the two types of models, we switch to a new SID at the beginning of each training stage.

Incorporating a stage identifier in the condition (text or class condition) across different training stages helps anchor the model's prior knowledge of the class learned in earlier stages. This approach effectively maps the unique distribution specific to the current stage to its corresponding stage identifier, minimizing the perturbation of the original function when switching stages.

\subsection{Automated Progressive Fine-tuning}

\begin{algorithm}[t]
\footnotesize
\color{newcolor}
\SetAlgoLined
\textbf{Input:}\\  
$|s|$:~total~training~steps;
$\tau$:~steps per stage;\\
Pre-trained parameters $\bm\omega$;\\
\For {$s \in [1,|s|]$}{
    \If{~~$s=N\tau,~~N\in\mathbb{N_{+}}$~~}{
    Calculate the zero-shot metrics of each candidate learnable sub-network;\\
    Search for the \textit{optimal learnable sub-network}~$\bm\psi^\star$ by \cref{eq:objectivezero_4};\\
    Unfreezing the \textit{optimal learnable sub-network}~~$\bm\psi\leftarrow\bm\psi^\star$;
    }{}
    Train the whole network, with $\bm\psi$ being the learnable part, over one batch of the training data.
}
\caption{AutoProg-\textit{Zero} for Fine-tuning}
\label{alg:autoprogzero}
\end{algorithm}

In this section, we focus on the automatic optimization of the unfreezing schedule $\Psi$ in Progressive Fine-tuning. We first formulate the multi-objective optimization problem of $\Psi$ on the task of diffusion models fine-tuning, then propose our solution, called AutoProg-\textit{Zero}, by generalizing AutoProg on fine-tuning and then introduce a novel zero-shot evaluation scheme for the unfreezing schedule $\Psi$.%

\subsubsection{Automated Progressive Fine-tuning by Zero-shot Metrics}\label{sec:subnetft}

Similar to the case for pre-training, the optimization of the unfreezing schedule $\Psi$ for efficient fine-tuning is a multi-objective bi-level optimization problem following \cref{eq:psi}. 
As optimization of \cref{eq:psi} contains optimization of network parameters $\bm\omega$, a full $|s|$ steps training is needed to get the optimal  $\bm\omega^\star$ for each unfreezing schedule $\Psi$. 
Similar to the case in AutoProg-\textit{One}, performing such extensive search over the higher level factor $\Psi$ in this bi-level optimization problem has complexity $\mathcal{O}(|\bm{\psi}|^{|k|}\cdot|s|)$.
We reduce the search cost to $\mathcal{O}(|\Lambda_k|\cdot|k|)$ by relaxing the original objective of unfreezing schedule search to progressively optimize sub-network architecture $\bm\psi^*_k$ in each stage $k$, following the relaxation in \cref{sec:subnet}.

\noindent\textbf{Limitation of One-shot Schedule Search.}
In AutoProg-\textit{One} for efficient pre-training, we optimize sub-network architecture $\bm\psi^*_k$ through their evaluation loss after one-shot training with Elastic Supernet. Through this, the search cost is further reduced to $\mathcal{O}(|k|)$ and then \textbf{\textit{zero}} through weight recycling, in \cref{sec:supernet}. However, this approximation is not suitable for fine-tuning and progressive unfreezing. In progressive unfreezing, all candidate learnable configurations have the same forward function and evaluation performance after one-shot training. As an alternative, we seek to estimate the future performance of different unfreezing schedules through analysis of the backward pass and gradients.
In sparse training and efficient Auto-ML algorithms, it is a common practice to estimate future ranking of models
with current parameters and their gradients~\cite{evci2020rigging,tanaka2020pruning}, or with parameters after a single step of gradient descent update~\cite{Liu2018DARTSDA,Cai2018ProxylessNASDN,Li2020DADADA}. These methods are not suitable for AutoProg-\textit{One} for progressive pre-training, as the network function is drastically changed and is not stable after growing.
In contrast, during progressive fine-tuning, the network function remains unchanged after unfreezing.
This approach enables the possibility of a \textit{zero-shot search} for the candidate unfreezing schedule.

\noindent\textbf{AutoProg-\textit{Zero}.}
Let $\mathcal{H}$ be the zero-shot evaluation metric to predict the final loss of each learnable sub-network. Denoting $\bm{\omega}^\star$ the zero-shot sub-network parameters directly inherited from the parameters of previous training, the optimal sub-network can be searched by:
\begin{equation}
\label{eq:objectivezero_2}
\begin{aligned}
    &\bm\psi_k^* = \mathop{\arg\min}_{\bm{\psi}_k\in\Lambda_k}\big\{\mathcal{H}\big(\bm{\omega}^\star(\bm{\psi}_k)\big), \mathcal{T}(\bm{\psi}_k)\big\},\\
\end{aligned}
\end{equation}
where $\Lambda_k$ denotes the unfreezing search space of the $k$-th stage. By directly performing zero-shot evaluation on the inherited parameters, the bi-level optimization problem is relaxed to a single-level one. The process of AutoProg-\textit{Zero} is shown in \cref{fig:autoprogzero} and \cref{alg:autoprogzero}.

Overall, by relaxing the original bi-level optimization problem in \cref{eq:objective} to \cref{eq:objectivezero_2}, we avoid the cost of optimizing the parameters $\bm\omega$ of the candidates. Thus, the search complexity is reduced drastically from $\mathcal{O}(|\bm{\psi}|^{|k|}\cdot|s|)$ to \textbf{\textit{zero}}.

\subsubsection{Zero-shot Proxy for Automated Progressive Learning}
During progressive unfreezing, all the candidate choices of unfreezing schedules have the same model function (forward pass). Therefore, their loss and other performance are the same. We can not use loss at the current step as a proxy for the loss of the target step at the end of the current training stage.
We design zero-shot metrics $\mathcal{H}$ to measure the trainability, convergence rate, and generalization capacity of candidate learnable sub-networks at the beginning of each stage to predict their performance after the training of this stage. By doing this, we estimate the future ranking of models with current parameters and their gradients. 

Suppose we train a diffusion model denoiser function $\bm{\epsilon}(\bm{x}, k)$ with parameters $\bm\omega$ using dataset $\{\bm{x}\}$. The forward diffusion process progressively perturbs a training sample $\bm{x}_0$ to a noisy version $\bm{x}_{k \in [0, 1]}$ by adding Gaussian noise. In the reverse process, diffusion model progressively denoises the noisy sample for $k$ steps from $\bm{x}_1$ to recover the original sample. At each denoising timestep $k$, the noise $\hat{\bm{\epsilon}}$ is predicted by a diffusion denoiser network $\bm{\epsilon}(\bm{x}, k)$. The optimization objective of this denoiser network is to minimize the mean square error loss:
\begin{equation}
      \mathcal{L}({\bm{\omega}} ; k) = \mathbb{E}_{\bm x_0,\bm x_k, \epsilon} [||\bm{\epsilon}_{\bm{\omega}}(\bm x_k, k) - \hat{\bm{\epsilon}}||^2].
\end{equation}

\noindent\textbf{Trainability by Condition Number of NTK.} The trainability of a neural network~\cite{chen2021neural,xiao2020disentangling} indicates how efficiently it can be optimized using gradient descent. Although large and complex networks are theoretically capable of representing intricate functions, they may not always be easily trainable through gradient descent. The Neural Tangent Kernel (NTK) has proven to be a valuable tool for analyzing the training dynamics of such networks~\cite{jacot2018neural}.

Let $\bm{x}$ be a batch of training samples for diffusion models at training step $s$. Consider the reverse diffusion process at the denoising timestep $k\in [0, 1]$. The noisy input used to predict the noise is $\bm{x}_k$.
In the case of progressive fine-tuning, we denote all the learnable parameters of a partially frozen neural network with $\bm\omega$ and only study the evolution of these parameters.
During the gradient descent training, the evolution of these parameters $\Delta{\bm{\omega}}_s$ and the output, \ie the predicted noise, $\Delta\bm{\epsilon}_s$ can be denoted as follows:
\begin{equation}
\begin{aligned}
    \label{eq:nn-gradient-descent-weights}
    \Delta{\bm{\omega}}_s &= 
    - \eta  {\nabla_{\bm\omega} \bm{\epsilon}_s(\bm{x}_k)}^\mathsf{T}
    \nabla_{\bm{\epsilon}_s(\bm{x}_k)} \mc L,\\
\end{aligned}
\end{equation}
\begin{equation}
\begin{aligned}
    \Delta{\bm{\epsilon}}_s(\bm{x}_k) &= {\nabla_{\bm\omega} \bm{\epsilon}_s(\bm{x}_k)}\, \Delta {\bm\omega}_s \\
    &= - \eta  \, \finntk_s (\bm{x}_k, \bm{x}_k)  \nabla_{\bm{\epsilon}_s(\bm{x}_k)} \mc L,
    \label{eq:nn-gradient-descent-outputs}
\end{aligned}  
\end{equation}
where $\eta$ denotes the learning rate and $\finntk_s (\bm{x}_k, \bm{x}_k) = {\nabla_{\bm\omega} \bm{\epsilon}_s(\bm{x}_k)}{\nabla_{\bm\omega} \bm{\epsilon}_s(\bm{x}_k)}^\mathsf{T}$ is the Neural Tangent Kernel (NTK) of partly frozen diffusion models at training step $s$. $\nabla_{\bm{\epsilon}_s(\bm{x}_k)} \mc L$ is the gradient of the loss function $\mc L$ with respect to the model’s output, the predicted noise $\bm{\epsilon}_s(\bm{x}_k)$.

The main result in~\cite{jacot2018neural} demonstrates that, in the infinite-width limit, the NTK converges to a deterministic kernel, denoted as $\infntk$, and remains constant throughout training. As a result, during gradient descent with an MSE loss, the expected outputs of an infinitely wide network, $\mu_s(\bm{x}_k) = \mathbb E[\bm{\epsilon}_i(\bm{x}_k)]$, evolve at training step $s$ as:
\begin{align}\label{eq:fc_ntk_recap_dynamics}
    \mu_s(\bm{x}_k) &= ({\Id} - e^{-\eta s \infntk})\hat{\bm{\epsilon}}.
\end{align}

As analyzed in \cite{xiao2020disentangling}, \cref{eq:fc_ntk_recap_dynamics} can be rewritten in terms of the spectrum of $\Theta$ as:
        \begin{equation}\label{eq:fc_ntk_dynamics_eigen}
        \tilde\mu_t(\bm{x}_k)_i = (\text{Id} - e^{-\eta s\lambda_i})\hat{\bm{\epsilon}}_i,
\end{equation}
where $\lambda_i$ are the eigenvalues of $\Theta$, and $\tilde\mu_t(\bm{x}_k)$ represents the mean prediction expressed in the eigenbasis of $\Theta$. By ordering the eigenvalues such that $\lambda_0 \geq \cdots \geq \lambda_m$, it has been suggested in \cite{lee2019wide} that the maximum feasible learning rate scales as $\eta \sim 2 / \lambda_0$. Substituting this scaling for $\eta$ into \cref{eq:fc_ntk_dynamics_eigen}, we observe that the smallest eigenvalue converges exponentially at a rate given by $1/\kappa$, where $\kappa =  \lambda_0 / \lambda_m$ is the condition number. 

Consequently, if the condition number of the NTK associated with a neural network diverges, the network becomes untrainable~\cite{chen2021neural,xiao2020disentangling}. Therefore, we use $\kappa$ as a zero-shot metric for trainability:
\begin{equation}
\begin{aligned}
        &\mathcal{H}_\kappa(\bm\psi) = \frac{\lambda_0}{\lambda_m},\\
    \text{where} ~~ \lambda_0 \geq \cdots &\geq \lambda_m ~~ \text{are the eigenvalues of} ~~\Theta.
\end{aligned}
\end{equation}

\noindent\textbf{Convergence Rate and Generalization Capacity via Gradient Analysis.}
The convergence rate is also an indicator of a neural network's trainability. As analyzed in \cite{li2023zico}, both the convergence rate and generalization capacity of a neural network can be effectively assessed through gradient analysis. First, after a certain number of training steps $s$, a network with a higher absolute mean of gradients across different training samples and a smaller standard deviation $\sigma$ of gradients will have a lower training loss, indicating a faster convergence rate at each step. Second, after the same number of training steps $s$, a network with a smaller standard deviation $\sigma$ of gradients will tend to have lower largest eigenvalues of the NTK $\Theta$, implying a flatter loss landscape and, consequently, better generalization~\cite{lewkowycz2020large}.

We use ZiCo~\cite{li2023zico} which jointly considers both absolute mean and standard deviation values of the gradients. Here, we generalize ZiCo to a partially frozen neural network $\bm\psi$ by only performing gradient analysis on learnable parameters of the network, while keeping the forward pass that is used to calculate the loss unchanged. Note that the original ZiCo metric is positively correlated with network performance. Here, we add a minus to ZiCo to make it positively correlated with loss, allowing us to minimize it effectively. Denote $l_{\bm\psi}$ the set of all the layers containing learnable parameters and $\bm\omega_{\bm\psi}$ the set of all the learnable parameters in a certain layer, we have our generalized ZiCo proxy:
\begin{equation}
    \mathcal{H}_\textit{ZiCo}(\bm\psi) = -\sum_{l\in l_{\bm\psi}}log\big(\sum_{\bm\omega\in\bm\omega_{\bm\psi}}\frac{\mathbb{E}[|\nabla_{\bm\omega}\mathcal{L}(\bm\omega; \bm{x})|]}{\sigma[|\nabla_{\bm\omega}\mathcal{L}(\bm\omega; \bm{x})|]}\big).
\end{equation}

To sum up, by using these two zero-shot proxies for schedule search in each stage $k$, \cref{eq:objectivezero_2} can be relaxed to:
\begin{equation}
\label{eq:objectivezero_3}
\begin{aligned}
    &\bm\psi_k^\star = \mathop{\arg\min}_{\bm{\psi}_k\in\Lambda_k}\big\{\mathcal{H}_\kappa(\bm\psi_k), \mathcal{H}_\textit{ZiCo}(\bm\psi_k), \mathcal{T}(\bm{\psi}_k)\big\}.\\
\end{aligned}
\end{equation}
In AutoProg-\textit{One}, we approximated the Pareto optimal solutions of a multi-objective optimization problem using a customized weighted product, as described in \cref{eq:psi_product}. However, this approach requires careful tuning of the hyperparameter $\alpha$. In AutoProg-\textit{Zero}, we eliminate the need for hyperparameters by adopting a ranked voting algorithm. Assuming that training efficiency and network performance are equally important, and that various zero-shot metrics hold equal significance in estimating network performance, we aggregate the rankings of these multiple objectives to obtain the final multi-objective ranking:
\begin{equation}
% \small
\label{eq:objectivezero_4}
\begin{aligned}
&\bm\psi_k^\star = \mathop{\arg\min}_{\bm{\psi}_k\in\Lambda_k}\textit{R}(\bm\psi_k),\\
\text{s.t.}~\textit{R}(\bm\psi_k) = \frac{1}{2}\textit{R}&(\mathcal{H}_\kappa(\bm\psi_k)) + \frac{1}{2}\textit{R}(\mathcal{H}_\textit{ZiCo}(\bm\psi_k)) + \textit{R}(\mathcal{T}(\bm{\psi}_k)),\\
\end{aligned}
\end{equation}

where $\textit{R}(\cdot)$ denotes the rank score. For example, the 1st, 2nd, 3rd, ... smallest candidates on each ballot receive 1, 2, 3, ... rank scores, and the candidate $\bm\psi_k$ with the smallest number of rank scores is selected as $\bm\psi_k^\star$.

}

\begin{figure*}[t]
    \centering
    \includegraphics[width=\linewidth]{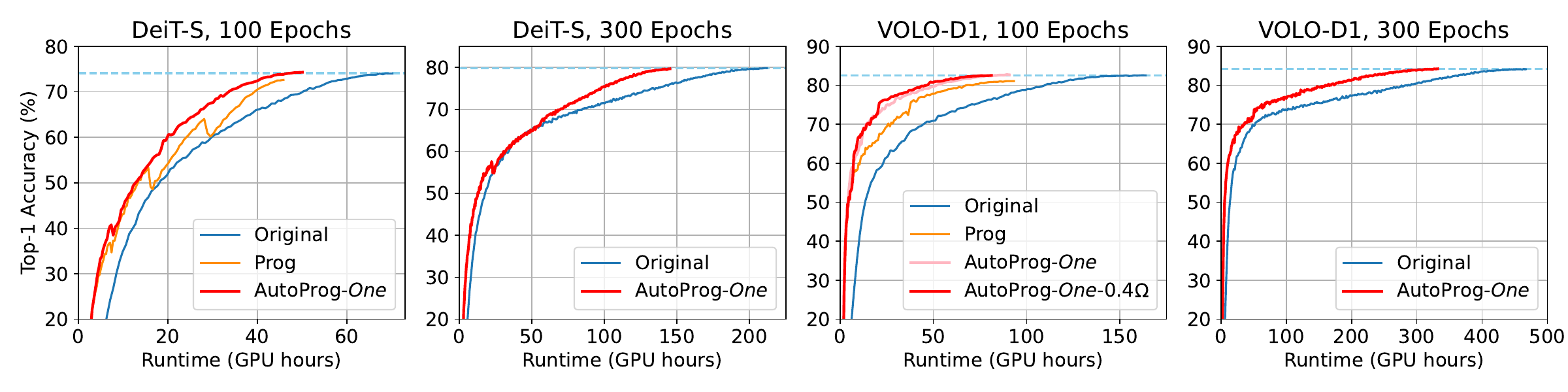}
    \caption{Evaluation accuracy of DeiT-S and VOLO-D1 during training with different learning schemes.%
    }
    \label{fig:learning_curve}
\end{figure*}

\section{Experiments}\label{sec:exp}

\subsection{Implementation Details}
\subsubsection{Implementation Details for Efficient Pre-training}
\noindent\textbf{Datasets.}
We evaluate our method on a large scale image classification dataset, ImageNet-1K~\cite{deng2009imagenet} and two widely used classification datasets, CIFAR-10 and CIFAR-100~\cite{Krizhevsky09cifar}, for transfer learning. ImageNet contains 1.2M train set images and 50K val set images in 1,000 classes. We use all the training data for progressive learning and supernet training, and use a 50K randomly sampled subset to calculate training loss for sub-network search.

\noindent\textbf{Architectures.}
We use two representative ViT architectures, DeiT~\cite{touvron2020deit} and VOLO~\cite{Yuan2021VOLOVO} to evaluate the proposed AutoProg.
Specifically, DeiT~\cite{touvron2020deit} is a representative standard ViT model; 
VOLO~\cite{Yuan2021VOLOVO} is a hybrid architecture comprised of \textit{outlook attention} blocks and transformer blocks.

\noindent\textbf{Training Details.}
For both architectures, we use the original training hyper-parameters, data augmentation and regularization techniques of their corresponding prototypes~\cite{touvron2020deit,Yuan2021VOLOVO}. Our experiments are conducted on NVIDIA 3090 GPUs.
As the acceleration achieved by our method is orthogonal to the acceleration of mixed precision training~\cite{micikevicius2018mixed}, we use it in both original baseline training and our progressive learning.

\noindent\textbf{Grow Space $\bm\Omega$.}
We use 4 stages for progressive learning. The initial scaling ratio $s_1$ is set to ${0.5}$ or $0.4$; the corresponding grow spaces are denoted by $0.5\Omega$ and $0.4\Omega$. By default, we use $0.5\Omega$ for our experiments, unless mentioned otherwise. The grow space of $n$ and $l$ are calculated by multiplying the value of the whole model with 4 equispaced scaling ratios $s \in \{0.5,0.67,0.83,1.0\}$, and we round the results to valid integer values. We use \textit{Prog} to denote our manual progressive baseline with \textit{uniform linear growth schedule} as described in Sec. \ref{sec:GS}.
\new{
\subsubsection{Implementation Details for Efficient Fine-tuning}
\noindent\textbf{Datasets.}
We evaluate our method on 7 downstream image generation tasks, including class-conditional generation on ArtBench-10~\cite{liao2022artbench}, CUB-200-2011~\cite{wah2011caltech}, Oxford Flowers~\cite{nilsback2008automated} and Stanford Cars~\cite{krause20133d}, text-to-image generation on CUB-200-2011~\cite{wah2011caltech} and Oxford Flowers~\cite{nilsback2008automated} and customized text-to-image generation on DreamBooth dataset~\cite{ruiz2023dreambooth}.
To use classification datasets without text annotation on text-to-image generation task, we constructed a standardized text prompt for training: ``a $<$class name$>$.''
For customized text-to-image generation, we use the first 4 images of the class ``\textsc{dog6}'' in DreamBooth dataset.

\noindent\textbf{Architectures.}
In our experiments, we employed DiT-XL/2\footnote{https://dl.fbaipublicfiles.com/DiT/models/DiT-XL-2-256x256.pt}, which achieved a remarkable FID score of 2.27 on the ImageNet 256×256 dataset after 7 million training iterations. For the text-to-image generation task using Stable Diffusion, we utilized the pre-trained Stable Diffusion\footnote{https://huggingface.co/runwayml/stable-diffusion-v1-5} model, known for its strong performance in generating high-quality images from textual descriptions. Additionally, for the DreamBooth framework, we followed the same experimental setup as DiffFit, using the Stable Diffusion\footnote{https://huggingface.co/CompVis/stable-diffusion-v1-4} model.

\noindent\textbf{Training Details.}
In our experiments with DiT, we adhered to the DiffFit settings, using a constant learning rate of 1e-4 for our proposed method. We configured the classifier-free guidance to 1.5 during evaluation and 4.0 for visualization, ensuring methodological consistency and enabling direct comparisons across different approaches. Our experiments were conducted on 8 A800 GPUs, utilizing a total batch size of 256 and 240K fine-tuning steps. The resolution for all datasets was uniformly resized to 256×256 pixels. FID scores were reported using 50 sampling steps.
For Stable Diffusion, we employed a constant learning rate of 1e-5, with the classifier-free guidance set to 3.0 for evaluation. Training was carried out on a single TITAN RTX GPU with a batch size of 32 over 32 epochs. The DDIM sampler was used, and FID scores were similarly reported after 50 sampling steps.
Following DreamBooth,  we set the learning rate to 5e-6. For DiffFit, we used the official codebase without any modifications. When using AutoProg-\textit{One} as a baseline for fine-tuning, we rank the learnable sub-networks by their average loss over the training process of the supernet, as all the candidates perform the same (have the same forward pass) after the training. %

\noindent\textbf{Grow Space $\bm\Omega$.}
We use 4 stages for progressive learning. The initial scaling ratio $s_1$ is set to ${0.25}$; the corresponding grow spaces are denoted by $0.25\Omega$. The grow space of $n$ and $l$ are calculated by multiplying the value of the whole model with 4 equispaced scaling ratios $s \in \{0.25,0.50,0.75,1.0\}$.
}

\begin{table}[t]
    \footnotesize
    \centering
    \setlength{\tabcolsep}{4pt}
    \begin{tabular}{l|l|c|c|c}
        \toprule
        Model  & \makecell[l]{Training\\scheme}  & \makecell{Speedup\\runtime}    & \makecell{Top-1\\(\%)} & \makecell{Top-1@288\\(\%)}\\
        \midrule
        \multicolumn{4}{l}{\textbf{\textit{100 epochs}}}\\
        \midrule
        \multirow{3}{*}{DeiT-S~\cite{touvron2020deit}}          & Original     & \na           & 74.1      & 74.6\\ %
        & Prog                                                     & \textcolor{gray}{+53.6\%} & 72.6      &  73.2\\ %
        & \cellcolor{Light}AutoProg-\textit{One}                                                 & \cellcolor{Light}+40.7\%                   & \cellcolor{Light}\textbf{74.4}      & \cellcolor{Light}\textbf{74.9}\\ %
        \arrayrulecolor{lightgray}\hline\arrayrulecolor{black}
        \multirow{4}{*}{VOLO-D1~\cite{Yuan2021VOLOVO}}          & Original     & \na           & 82.6      &83.0\\ %
        & Prog                                                     & \textcolor{gray}{+60.9\%} & 81.7      &82.1\\ %
        & \cellcolor{Light}AutoProg-\textit{One}                                     &\cellcolor{Light} +65.6\%                   & \cellcolor{Light}\textbf{82.8}      &\cellcolor{Light}\textbf{83.2}\\ %
        & \cellcolor{Light}AutoProg-\textit{One}-0.4$\Omega$                                     & \cellcolor{Light}\textbf{+85.1\%}          &\cellcolor{Light} 82.7      & \cellcolor{Light}83.1 \\ %
        \arrayrulecolor{lightgray}\hline\arrayrulecolor{black}
        \multirow{3}{*}{VOLO-D2~\cite{Yuan2021VOLOVO}}          & Original     & \na                       & 83.6      & 84.1\\ %
        & Prog                                                     & \textcolor{gray}{+54.4\%} & 82.9      & 83.3\\ %
        & \cellcolor{Light}AutoProg-\textit{One}                                                 & \cellcolor{Light}+45.3\%                   &\cellcolor{Light} \textbf{83.8}      & \cellcolor{Light}\textbf{84.2}\\ %
        \midrule
        \multicolumn{4}{l}{\textbf{\textit{300 epochs}}}\\
        \midrule
        \multirow{2}{*}{DeiT-Tiny~\cite{touvron2020deit}}       &  Original    & \na           & 72.2      & 72.9\\ %
        &\cellcolor{Light}AutoProg-\textit{One}                                                 &    \cellcolor{Light}+51.2\%                       & \cellcolor{Light} \textbf{72.4}  & \cellcolor{Light}\textbf{73.0}  \\ %
        \arrayrulecolor{lightgray}\hline\arrayrulecolor{black}
        \multirow{2}{*}{DeiT-S~\cite{touvron2020deit}}          &  Original    & \na           & 79.8      &  80.1\\ %
        & \cellcolor{Light}AutoProg-\textit{One}                                                 & \cellcolor{Light}+42.0\%                   &\cellcolor{Light}79.8      & \cellcolor{Light} 80.1 \\ %
        \arrayrulecolor{lightgray}\hline\arrayrulecolor{black}
        \multirow{2}{*}{VOLO-D1~\cite{Yuan2021VOLOVO}}          & Original     & \na           & 84.2      & 84.4 \\ %
        & \cellcolor{Light}AutoProg-\textit{One}                                                 & \cellcolor{Light}+48.9\%                   & \cellcolor{Light}\textbf{84.3}      &\cellcolor{Light}\textbf{84.6}\\ %
        \arrayrulecolor{lightgray}\hline\arrayrulecolor{black}
        \multirow{2}{*}{VOLO-D2~\cite{Yuan2021VOLOVO}}          &Original      & \na           & 85.2      & 85.1\\
        &\cellcolor{Light}AutoProg-\textit{One}                                                 & \cellcolor{Light}+42.7\%                   &\cellcolor{Light} 85.2     & \cellcolor{Light} \textbf{85.2}\\
        \bottomrule
    \end{tabular}%
    \caption{\textbf{Efficient pre-training results on image classification on ImageNet.}
    Accelerations that cause accuracy drop are marked with \textcolor{gray}{gray}. Best results are marked with \textbf{Bold}; our method or default settings are highlighted in \colorbox{Light}{purple}. Top-1@288 denotes Top-1 Accuracy when directly testing on 288$\times$288 input size, \textit{without} finetuning.
    Please refer to Fig.~\ref{fig:learning_curve} for the accuracy during training.
    }
    \label{tab:imagenet}
\end{table}
\begin{table}[t]
\setlength{\tabcolsep}{9pt}
\footnotesize
    \centering
    \begin{tabular}{lccc}
    \toprule
    Pretrain   & Speedup & CIFAR-10  & CIFAR-100\\
    \midrule
    Original    &\na        &99.0   &89.5  \\
    \rowcolor{Light}AutoProg-\textit{One}    &\bf{48.9}\%     &\bf{99.0}   &\bf{89.7}   \\
    \bottomrule
    \end{tabular}%
    \caption{\textbf{Transfer learning results of efficiently pre-trained DeiT-S on CIFAR datasets.} The evaluation metric is Top-1 accuracy (\%).}
    \label{tab:transfer}
\end{table}

\subsection{Efficient Pre-training}
\subsubsection{Efficient Pre-training on ImageNet}\label{sec:exp_imagenet}
We first validate the effectiveness of AutoProg-\textit{One} for efficient pre-training on ImageNet. As shown in \cref{tab:imagenet}, AutoProg-\textit{One} consistently achieves remarkable efficient training results on diverse ViT architectures and training schedules.
\textbf{First}, our AutoProg-\textit{One} achieves significant training acceleration over the regular training scheme with no performance drop. Generally, AutoProg-\textit{One} speeds up ViTs training by more than 45\% despite changes on training epochs and network architectures. In particular, VOLO-D1 trained with \textit{AutoProg} 0.4$\Omega$ achieves \textbf{85.1\%} training acceleration, and even slightly improves the accuracy (+0.1\%).
\textbf{Second}, AutoProg-\textit{One} outperforms the manual baseline, the uniform linear growing (Prog), by a large margin. For instance, Prog scheme causes severe performance degradation on DeiT-S. AutoProg-\textit{One} improves over Prog scheme on DeiT-S by \textbf{1.7\%} on accuracy, successfully eliminating the performance gap by automatically choosing the proper growth schedule.
\textbf{Third}, as progressive learning uses smaller input size during training, one may question its generalization capability on larger input sizes. We answer this by directly testing the models trained with AutoProg-\textit{One} on 288$\times$288 input size. The results justify that models trained with AutoProg-\textit{One} have comparable generalization ability on larger input sizes to original models. Remarkably, VOLO-D1 trained for 300 epochs with AutoProg-\textit{One} reaches \textbf{84.6\%} Top-1 accuracy when testing on 288$\times$288 input size, with \textbf{48.9\%} faster training.

The learning curves (\ie, evaluation accuracy during training) of DeiT-S and VOLO-D1 with different training schemes are shown in \cref{fig:learning_curve}.
AutoProg-\textit{One} clearly accelerates the training progress of these two models. Interestingly, DeiT-S (100 epochs) trained with manual Prog scheme presents \textit{sharp fluctuations} after growth, while AutoProg-\textit{One} successfully addresses this issue and eventually reaches higher accuracy by choosing proper growth schedule.
\subsubsection{Transfer Learning of Efficiently Pre-trained Models}

To further evaluate the transfer ability of ViTs trained with AutoProg, we conduct transfer learning on CIFAR-10 and CIFAR-100 datasets.
We use the DeiT-S model that is pretrained with AutoProg-\textit{One} on ImageNet for finetuning on CIFAR datasets, following the procedure in~\cite{touvron2020deit}. We compare with its counterpart pretrained with the ordinary training scheme.
The results are summarized in \cref{tab:transfer}. While AutoProg-\textit{One} largely saves training time, it achieves competitive transfer learning results. This proves that AutoProg-\textit{One} acceleration on ImageNet pretraining does not harm the transfer ability of ViTs on CIFAR datasets. %

\begin{figure}[t]
    \vspace{-5pt}
    \centering
    \includegraphics[width=0.8\linewidth]{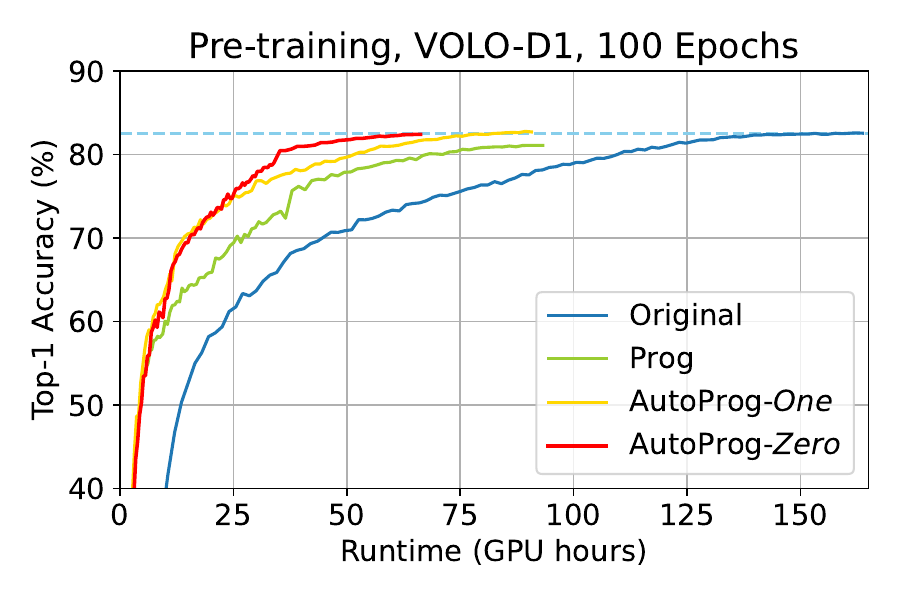}
    \vspace{-5pt}
    \caption{\new{\textbf{AutoProg-\textit{Zero} on pre-training.} Evaluation accuracy of VOLO-D1 during training with different learning schemes.%
    }}
    \label{fig:learning_curve_zero}
\end{figure}

\begin{table}[t]
    \color{newcolor}
    \setlength{\tabcolsep}{8pt}
    \footnotesize
    \centering
    \begin{tabular}{l|c|c}
        \toprule
        Method  & \makecell{Top-1 Acc} &  Speedup runtime \\
        \midrule
        Original & 82.6  & -\\
        \midrule
        Prog & 81.7 & +60.9\%\\
        \rowcolor{Light}AutoProg-\textit{One} & \textbf{82.7} & +85.1\%\\

        \rowcolor{Light}AutoProg-\textit{Zero}  & 82.5 & \textbf{+108.3\%}\\

        \bottomrule
    \end{tabular}
    \caption{\new{\textbf{Efficient pre-training results of AutoProg-\textit{Zero} on image classification on ImageNet.} We compare the Top-1 Accuracy (\%) by pre-training VOLO-D1 for 100 epochs. Please refer to Fig.~\ref{fig:learning_curve_zero} for the accuracy during training.}}
    \label{tab:zero_pretrain}
\end{table}

\begin{figure*}[t]
    \centering
    \includegraphics[width=\linewidth]{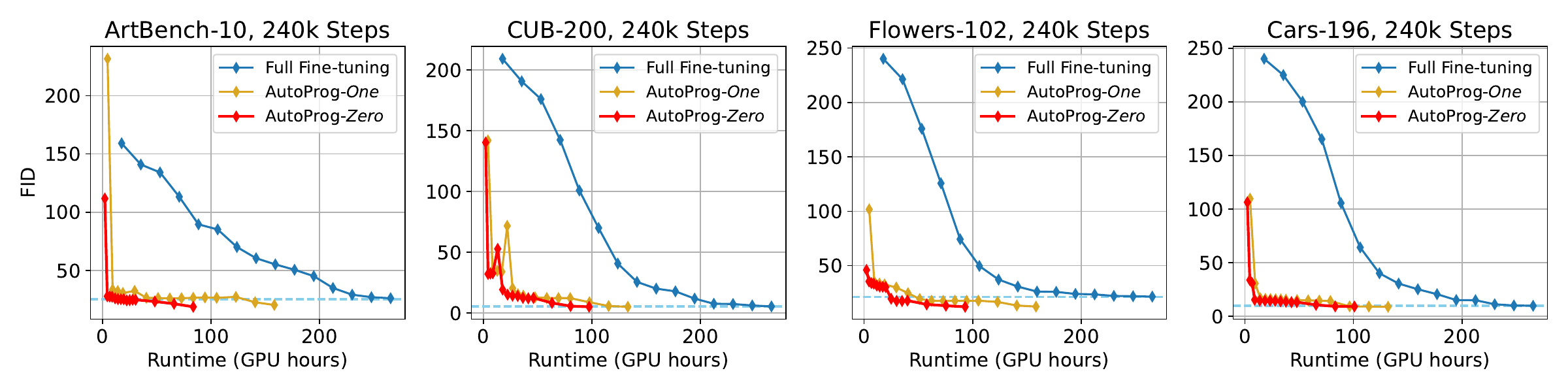}
    \caption{\textbf{FID of different fine-tuning methods every 15K iterations on four downstream datasets.} Our two methods both converge at a remarkable speed at the beginning of fine-tuning. AutoProg-\textit{Zero} achieves the best speedup and performance.}
    \label{fig:fid_curve}
\end{figure*}

\subsubsection{AutoProg-\textit{Zero} on Efficient Pre-training.}

Our AutoProg-\textit{Zero} is designed specifically for efficient fine-tuning. Here, we explore its generalization performance on the efficient pre-training task on ImageNet using the VOLO-D1 model. For a fair comparison, AutoProg-\textit{Zero}, AutoProg-\textit{One}, and Prog use the same grow space, 0.5$\Omega$. Note that we disable the SID scheme when applying AutoProg-\textit{Zero} to ViT pre-training. We compare AutoProg-\textit{Zero} with our default method for pre-training, AutoProg-\textit{One}, and our manually designed baseline method, Prog. As shown in Tab.~\ref{tab:zero_pretrain}, our default method AutoProg-\textit{One} achieves the best performance on this task with substantial speedup in total runtime of +85.1\%. Remarkably, AutoProg-\textit{Zero} achieves an even higher speedup of +108.3\%, with comparable accuracy. Such advantage of AutoProg-\textit{Zero} is visualized in Fig.~\ref{fig:learning_curve_zero} by comparing the accuracy of networks trained by different methods regarding the total runtime during training. These results indicate that AutoProg-\textit{Zero} has the potential to generalize beyond fine-tuning to other training scenarios, such as ImageNet pre-training. This strong generalization ability may be attributed to the effectiveness of zero-shot proxies, which remain reliable even on randomly initialized networks.

\begin{table}[t]
    \color{newcolor}
    \footnotesize
    \centering
    \setlength{\tabcolsep}{3pt}
    \begin{tabular}{l|c|c|c|c|c}
        \toprule
        \diagbox{Method}{Dataset}  & \makecell{Oxford\\Flowers}  & ArtBench & \makecell{CUB\\Bird} & \makecell{Stanford\\Cars} & \makecell{Average\\Runtime}\\
        \midrule
        Full Fine-tuning & 21.05  & 25.31 & 5.68 & 9.79 & 1×\\
        Adapt-Parallel~\cite{chen2022adaptformer} & 21.24  & 38.43 & 7.73 & 10.73 & 0.47×\\
        Adapt-Sequential & 21.36  & 35.04 & 7.00 & 10.45 & 0.43×\\
        BitFit~\cite{zaken2021bitfit} & 20.31   & 24.53 & 8.81 & 10.64 & 0.45×\\
        VPT-Deep~\cite{jia2022visual} & 25.59  & 40.74 & 17.29 & 22.12 & 0.50×\\
        LoRA-R8~\cite{hu2021lora} & 164.13  & 80.99 & 56.03 & 76.24 & 0.63×\\
        LoRA-R16 & 161.68 & 80.72 & 58.25 & 75.35 & 0.68×\\
        DiffFit~\cite{xie2023difffit} & 20.18  & 20.87 & 5.48 & 9.90 & 0.49×\\
        \midrule
        AutoProg-\textit{One} & 12.30  & 19.48 & 5.31 & 8.80 & 0.58×\\
        \rowcolor{Light}AutoProg-\textit{Zero}  & \textbf{12.19}  & \textbf{18.43} & \textbf{5.20} & \textbf{8.79} & \textbf{0.39×}\\

        \bottomrule
    \end{tabular}
    \caption{\new{\textbf{Efficient fine-tuning results on class-conditional image generation.} We compare FID using DiT-XL-2 pre-trained on ImageNet 256×256.
    Please refer to Fig.~\ref{fig:fid_curve} for the accuracy during training.%
    }}
    \label{tab:dit_fid_results}
\end{table}

\begin{table}[t]
    \color{newcolor}
\setlength{\tabcolsep}{3pt}
    \footnotesize
    \centering
    \begin{tabular}{l|cc|cc|c}
        \toprule
        \multirow{2}{*}{\diagbox{Method}{Dataset}}  &\multicolumn{2}{c|}{\makecell{Oxford\\Flowers}} & \multicolumn{2}{c|}{CUB-Bird} &\multirow{2}{*}{\makecell{Average\\Runtime}}\\
        & FID$\downarrow$ & CLIP-T$\uparrow$ & FID$\downarrow$ & CLIP-T$\uparrow$\\
        \midrule
        Zero-Shot Transfer & 223.93 & 0.224 & 157.92 & 0.230 & -\\
        Full Fine-tuning  & 35.21 &0.324& 9.32 &0.322& 1×\\
        DiffFit~\cite{xie2023difffit} & 72.28 &0.286& 12.04 &0.311& 1.20×\\

        \midrule
        Prog & 32.16 &0.324& 9.11 &0.326& 0.51×\\
        AutoProg-\textit{One} & 32.71 &0.325& 8.92 & 0.326 & 0.45×\\
        \rowcolor{Light}AutoProg-\textit{Zero} & \textbf{31.91} &\textbf{0.328}& \textbf{8.74} &\textbf{0.327}& \textbf{0.39×}\\

        \bottomrule
    \end{tabular}
    \caption{\textbf{Efficient fine-tuning results on text-to-image generation with Stable Diffusion.}%
    }
    \label{tab:sd_fid_results}
\end{table}

\begin{table}[t]
    \color{newcolor}

    \footnotesize
    \centering
    \begin{tabular}{l|c|c|c|c}
        \toprule
        Method  & \makecell{DINO$\uparrow$}  &  \makecell{CLIP-T$\uparrow$\\Prompt-A}  & \makecell{CLIP-T$\uparrow$\\Prompt-B} & Runtime \\
        \midrule
        Original & 0.849 & 0.214 & 0.253 & 1×\\
        DiffFit~\cite{xie2023difffit} & 0.841 & 0.195 & 0.225 & 0.79×\\
        
        \midrule
        Prog & 0.857 & 0.236 & 0.283 & 0.44×\\
        AutoProg-\textit{One} & 0.858 & 0.251 & 0.280 & 0.39×\\
        \rowcolor{Light}AutoProg-\textit{Zero}  & \textbf{0.874}& \textbf{0.280} & \textbf{0.303}& \textbf{0.35×}\\

        \bottomrule
    \end{tabular}
    \caption{\textbf{Efficient fine-tuning results on customized text-to-image generation.} We compare different fine-tuning methods using DreamBooth with Stable Diffusion.
    Prompt-A: a photo of [V] dog sleeping under a tree. Prompt-B: a photo of [V] dog on the beach.
    }
    \label{tab:clip_results}
\end{table}

\new{
\subsection{Efficient Fine-tuning}

\subsubsection{Efficient Fine-tuning on Class-conditional Image Generation}
We perform efficient fine-tuning on class-conditional image generation with DiT.
As shown in \cref{tab:dit_fid_results}, AutoProg-\textit{Zero} significantly reduces the average fine-tuning time to just 39\% of the time required by full fine-tuning, speedup the runtime by \textbf{2.56×}\footnote{We use \textit{reduced time} and \textit{speedup} interchangeably, where $\frac{1}{n}$ reduced time is equivalent to $n$× speedup, or + [($n$-1)×100] \% speedup.} while consistently achieving the lowest FID scores across datasets. Notably, on the Oxford Flowers dataset, AutoProg-\textit{Zero} improves the FID to 12.19, achieving superior generative performance. As illustrated in \cref{fig:art_compare}, we visually compare the generation results of AutoProg-\textit{One}, AutoProg-\textit{Zero}, and Inadequate Fine-tuning on the Artbench dataset. The comparison reveals that, under similar total runtime, fine-tuning with AutoProg-\textit{Zero} generates local details and artistic features better.

\begin{figure*}[t]
    \centering
    \includegraphics[width=0.85\linewidth]{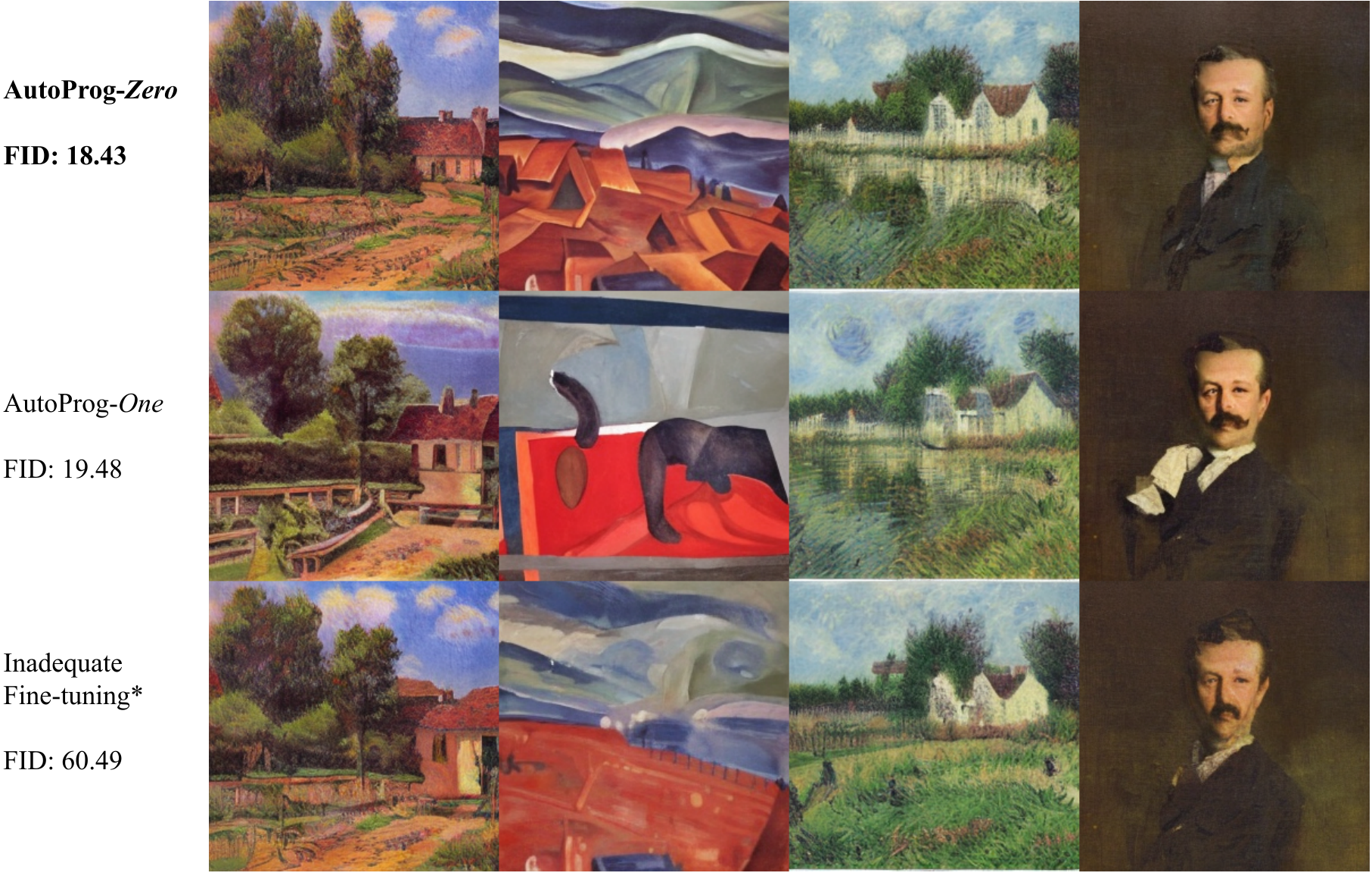}~~~~~~~~~~~
    \caption{Comparison of the generation results of different fine-tuning methods on the ArtBench dataset with DiT-XL/2. Inadequate Fine-tuning* represents the incomplete fine-tuning results under the same training time as AutoProg-\textit{Zero}.}
    \label{fig:art_compare}
\end{figure*}

\begin{figure*}[t]
    \centering
    \includegraphics[width=0.95\linewidth]{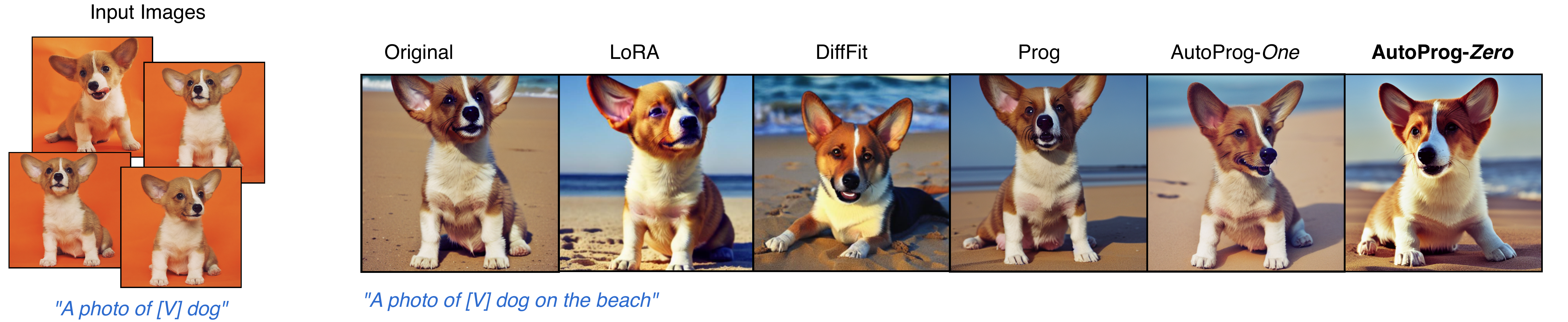}~~~
    \caption{Comparison of generated images of different fine-tuning method using DreamBooth with Stable Diffusion.\protect\footnotemark
    }
    \label{fig:dreambooth}
\end{figure*}

\subsubsection{Efficient Fine-tuning on Text-to-Image Generation}
We selected Stable Diffusion for our text-to-image fine-tuning task. As detailed in \cref{tab:sd_fid_results}, AutoProg-\textit{Zero} outperformed other fine-tuning methods, achieving superior CLIP scores and FID results on both the CUB and Flowers datasets. Notably, it also significantly reduced the time required for fine-tuning to 0.39×, achieving a speedup of \textbf{2.56×}. In contrast, DiffFiT exhibited sub-optimal performance, suggesting that the complexity of text-to-image generation may necessitate a more comprehensive approach, as fine-tuning only a small subset of model parameters may be insufficient to capture the intricate features required for this task.

\subsubsection{Efficient Fine-tuning on Customized Text-to-Image Generation}
We use DreamBooth on Stable Diffusion for customized text-to-image generation.
For the fidelity metrics, we use the DINO score~\cite{ruiz2023dreambooth} for image fidelity and the CLIP-T score for text fidelity. We choose DINO score~\cite{ruiz2023dreambooth} for image fidelity because DINO~\cite{caron2021emerging} is a better metric for subject-driven image generation due to its ability to distinguish unique features of a subject or image. For text fidelity, we use the CLIP-T metric that measures the fidelity of the generated images to the textual prompts.

\cref{tab:clip_results} presents the performance of different fine-tuning methods on DreamBooth. AutoProg-\textit{Zero} demonstrates a remarkable speedup of 2.86× (0.35× total runtime), with superior performance in generating complex semantic content, producing images more accurately reflecting text descriptions. As shown in \cref{fig:dreambooth}, the generation results highlight the effectiveness of AutoProg-\textit{Zero}. Notably, among the generated images, only original fine-tuning and AutoProg-\textit{Zero} properly generate the details of white foam on the waves. The subject fidelity and aesthetic appeal of AutoProg-\textit{Zero} are also better than other methods.

\begin{table}[t]
\footnotesize
    \centering
    \begin{tabular}{l|c|c}
    \toprule
    Growth Op. $\bm\zeta$                           & Top-1@Growth (\%)   & Top-1 (\%)  \\
    \midrule
    Baseline                                        &\na            & 82.53 \\
    \midrule
    RandInit~\cite{Simonyan2015VeryDC}              & 60.61         & 80.02 \\
    Stacking~\cite{Gong2019EfficientTO}              & 61.50         & 81.55 \\
    \rowcolor{Light}Interpolation~\cite{chang2018multi,Dong2020TowardsAR}& \textbf{61.53}& \textbf{81.78} \\
    \midrule
    Identity~\cite{Chen2016Net2NetAL,Wei2016NetworkM} & 61.04         & 79.32 \\
    \rowcolor{Light}MoGrow                          &\textbf{61.65} & \textbf{81.90} \\

    \bottomrule
    \end{tabular}%
    \caption{Ablation analysis of depth growth operator $\bm\zeta$ with the Prog learning scheme. Top-1@Growth denotes the accuracy after training for the first epoch of the second stage.%
    }
    \label{tab:ablation_growth}
\end{table}

\begin{table}[t]
\footnotesize
    \centering
    \begin{tabular}{l|c|c}
    \toprule
    Method   & Top-1@Growth (\%)  & Top-1 (\%)  \\
    \midrule
    AutoProg-\textit{One} w/o MoGrow                             & 59.41                 & 82.6 \\
    \rowcolor{Light}AutoProg-\textit{One} w/ MoGrow                              & \textbf{62.14}        & \textbf{82.8}\\
    \bottomrule
    \end{tabular}%
    \caption{Ablation analysis of \textit{MoGrow} in AutoProg-\textit{One} on VOLO-D1. Top-1@Growth denotes the accuracy of the supernet after training for the first epoch of the second stage.}
    \label{tab:ablation_MoGrow}
\end{table}

\begin{table}[t]
\footnotesize
    \centering
    \begin{tabular}{l|c|c}
    \toprule
    Method  & Speedup   & Top-1 Acc. (\%)  \\
    \midrule
    w/o recycling   & 53.3\%       &   82.8    \\
    \rowcolor{Light}w/ recycling    &\textbf{65.6\%}    & \textbf{82.8}\\
    \bottomrule
    \end{tabular}
    \caption{Ablation analysis of \textit{weight recycling} in AutoProg-\textit{One} on VOLO-D1.}
    \label{tab:ablation_weightRec}
\end{table}

\begin{table}[t]
    \color{newcolor}
    \footnotesize
    \centering
    \begin{tabular}{l|c|c|c}
        \toprule
        Method  & SID & Food & Runtime \\

        \midrule
        AutoProg-\textit{One} & - & 8.65 & 0.62×\\
        AutoProg-\textit{Zero} & \xmark & 8.61 & \textbf{0.39×}\\
        \rowcolor{Light}AutoProg-\textit{Zero}  & \cmark & \textbf{7.70} & \textbf{0.39×}\\
        \bottomrule
    \end{tabular}
    \caption{Ablation analysis of SID (embedding) in our AutoProg-\textit{Zero} with DiT on Food.}
    \label{tab:dit_autoprog_ablation}
\end{table}

\begin{table}[t]
    \color{newcolor}

    \footnotesize
    \centering
    \begin{tabular}{l|c|c|c}
        \toprule
        Method  & SID & CUB & Runtime \\

        \midrule
        AutoProg-\textit{One} & - & 8.92 & 0.39×\\
        AutoProg-\textit{Zero} & \xmark & 8.87 & \textbf{0.34×}\\
        \rowcolor{Light}AutoProg-\textit{Zero}  & \cmark & \textbf{8.74} & \textbf{0.34×}\\

        \bottomrule
    \end{tabular}
    \caption{Ablation analysis of SID (text) in our AutoProg-\textit{Zero} with Stable Diffusion on CUB.}
    \label{tab:sd_autoprog_ablation}
\end{table}

}

\footnotetext{Generated images of LoRA and DiffFit is taken from~\cite{xie2023difffit}.}

\subsection{Ablation Studies}\label{sec:exp_ablation}
\subsubsection{Ablation Studies on AutoProg-\textit{One} for Pre-training}
\noindent\textbf{Growth Operator $\bm\zeta$.}
We first compare the three growth operators mentioned in \cref{sec:GO}, \ie, \textit{RandInit}~\cite{Simonyan2015VeryDC}, \textit{Stacking}~\cite{Gong2019EfficientTO} and \textit{Interpolation}~\cite{chang2018multi,Dong2020TowardsAR}, by using them with manual Prog scheme on VOLO-D1. As shown in \cref{tab:ablation_growth}, \textit{Interpolation} growth achieves the best accuracy both after the first growth and in the final. 

Then, we compare two growth operators build upon \textit{Interpolation} scheme, our proposed MoGrow, and Identity, which is a function-preserving~\cite{Chen2016Net2NetAL,Wei2016NetworkM} operator that can be achieved by Interpolation + ReZero~\cite{Bachlechner2020ReZeroIA}.
Specifically, ReZero uses a zero-initialized, learnable scalar to scale the residual modules in networks. %
Using this technique on newly added layers can assure the original network function is preserved. The results are shown in \cref{tab:ablation_growth}. Contrary to expectations, we observe that Identity growth largely \textit{reduces} the Top-1 accuracy of VOLO-D1 (-3.21\%), probably because the network convergence is slowed down by the zero-initialized scalar; besides, the global minimum of the original function could be a local minimum in the new network, which hinders the optimization. On this inferior growth schedule, our MoGrow still improves over Interpolation by 0.15\%, effectively reducing its performance gap.

Previous comparisons are based on the Prog scheme. Moreover, we also analyze the effect of MoGrow on AutoProg. The results are shown in \cref{tab:ablation_MoGrow}. We observe that MoGrow largely improves the performance of the supernet by \textbf{2.73\%}. It also increases the final training accuracy by 0.2\%, proving the effectiveness of MoGrow in AutoProg.

\noindent\textbf{Weight Recycling.} We further study the effect of weight recycling by training VOLO-D1 using AutoProg. As shown in \cref{tab:ablation_weightRec}, by recycling the weights of the supernet, AutoProg-\textit{One} can achieve 12.3\% more speedup. Also, benefiting from the synergy effect in weight-nesting~\cite{yu2019slimmable}, weight recycling scheme does not cause accuracy drop. These results prove the effectiveness of weight recycling.

\new{
\subsubsection{Ablation Studies on AutoProg-\textit{Zero} for Fine-tuning}
\noindent\textbf{Unique Stage Identifier (SID).} We investigated the impact of the SID component of AutoProg-\textit{Zero} on Stable Diffusion and DiT. As shown in \cref{tab:dit_autoprog_ablation}, when applied to the DiT on the Foods~\cite{bossard2014food} dataset, the addition of the SID component to AutoProg-\textit{Zero} resulted in improved FID scores. Importantly, this enhancement did not compromise the speed of AutoProg-\textit{Zero}. Similarly, as detailed in \cref{tab:sd_autoprog_ablation}, we conducted the same experiment using Stable Diffusion on the CUB dataset, where AutoProg-\textit{Zero} with the SID component again achieved superior results. These findings suggest that the SID component effectively bridges the gap between different stages of model unfreezing, facilitating better convergence and leading to enhanced overall performance.

\noindent\textbf{Zero-shot search.} We compared the performance of AutoProg-\textit{One} with AutoProg-\textit{Zero} without the SID component, as presented in \cref{tab:dit_autoprog_ablation} and \cref{tab:sd_autoprog_ablation}. Our experiments on both Stable Diffusion and DiT reveal that AutoProg-\textit{Zero} consistently achieved superior FID scores. Additionally, AutoProg-\textit{Zero}'s ability to select the most suitable network architecture at each stage of training significantly reduced the overall fine-tuning time, \eg from 0.62× to 0.39× on DiT, further demonstrating the efficiency and effectiveness of zero-shot search.

\section{Conclusion and Discussion}
In this paper, we take a practical step towards sustainable deep learning by generalizing and automating progressive learning for LVMs. We have developed an Advanced AutoProg framework to improve the training efficiency of various learning scenarios of LVMs. Firstly, we present AutoProg-\textit{One}, featuring \textit{MoGrow} and one-shot search of the growth schedule, for efficient pre-training of ViTs.
Secondly, we introduce AutoProg-\textit{Zero}, a novel zero-shot automated progressive learning method for efficient fine-tuning of diffusion models, along with \textit{SID} to bridge the gap between different stages of model unfreezing.
Our AutoProg has achieved consistent pre-training and fine-tuning speedup on different LVMs without sacrificing performance. AutoProg-\textit{One} speedup pre-training of ViTs by up to 1.85× while maintaining comparable performance to traditional pre-training. AutoProg-\textit{Zero} achieves remarkable speedup on diffusion model fine-tuning by up to 2.86× with lossless generative performance. Ablation studies have proved the effectiveness of each component of AutoProg.

\noindent\textbf{Social Impact and Limitations.}
When network training becomes more efficient, it is also more available and less subject to regularization, which may result in a proliferation of models with harmful biases or intended uses.
In this work, we achieve inspiring results with automated progressive learning on LVMs. However, large scale training of large language models (LLMs) and other fields can not directly benefit from it. We encourage future works to develop automated progressive learning for efficient training in broader applications.

}

\ifCLASSOPTIONcaptionsoff
  \newpage
\fi

\bibliographystyle{IEEEtran}
\bibliography{IEEEabrv,egbib}

\begin{thebibliography}{100}
\providecommand{\url}[1]{#1}
\csname url@samestyle\endcsname
\providecommand{\newblock}{\relax}
\providecommand{\bibinfo}[2]{#2}
\providecommand{\BIBentrySTDinterwordspacing}{\spaceskip=0pt\relax}
\providecommand{\BIBentryALTinterwordstretchfactor}{4}
\providecommand{\BIBentryALTinterwordspacing}{\spaceskip=\fontdimen2\font plus
\BIBentryALTinterwordstretchfactor\fontdimen3\font minus \fontdimen4\font\relax}
\providecommand{\BIBforeignlanguage}[2]{{%
\expandafter\ifx\csname l@#1\endcsname\relax
\typeout{** WARNING: IEEEtran.bst: No hyphenation pattern has been}%
\typeout{** loaded for the language `#1'. Using the pattern for}%
\typeout{** the default language instead.}%
\else
\language=\csname l@#1\endcsname
\fi
#2}}
\providecommand{\BIBdecl}{\relax}
\BIBdecl

\bibitem{touvron2020deit}
H.~Touvron, M.~Cord, M.~Douze, F.~Massa, A.~Sablayrolles, and H.~J{\'e}gou, ``Training data-efficient image transformers \& distillation through attention,'' in \emph{ICML}, 2021.

\bibitem{liu2021swin}
Z.~Liu, Y.~Lin, Y.~Cao, H.~Hu, Y.~Wei, Z.~Zhang, S.~Lin, and B.~Guo, ``Swin transformer: Hierarchical vision transformer using shifted windows,'' in \emph{ICCV}, 2021.

\bibitem{dosovitskiy2021image}
A.~Dosovitskiy, L.~Beyer, A.~Kolesnikov, D.~Weissenborn, X.~Zhai, T.~Unterthiner, M.~Dehghani, M.~Minderer, G.~Heigold, S.~Gelly, J.~Uszkoreit, and N.~Houlsby, ``An image is worth 16x16 words: Transformers for image recognition at scale,'' in \emph{ICLR}, 2021.

\bibitem{he2016deep}
K.~He, X.~Zhang, S.~Ren, and J.~Sun, ``Deep residual learning for image recognition,'' in \emph{CVPR}, 2016.

\bibitem{zhai2021scaling}
X.~Zhai, A.~Kolesnikov, N.~Houlsby, and L.~Beyer, ``Scaling vision transformers,'' \emph{arXiv preprint arXiv:2106.04560}, 2021.

\bibitem{dai2021coatnet}
Z.~Dai, H.~Liu, Q.~V. Le, and M.~Tan, ``Coatnet: Marrying convolution and attention for all data sizes,'' in \emph{NeurIPS}, 2021.

\bibitem{peebles2023scalable}
W.~Peebles and S.~Xie, ``Scalable diffusion models with transformers,'' in \emph{ICCV}, 2023, pp. 4195--4205.

\bibitem{xie2023difffit}
E.~Xie, L.~Yao, H.~Shi, Z.~Liu, D.~Zhou, Z.~Liu, J.~Li, and Z.~Li, ``Difffit: Unlocking transferability of large diffusion models via simple parameter-efficient fine-tuning,'' in \emph{ICCV}, 2023, pp. 4230--4239.

\bibitem{Yuan2021VOLOVO}
L.~Yuan, Q.~Hou, Z.~Jiang, J.~Feng, and S.~Yan, ``{VOLO}: Vision outlooker for visual recognition,'' \emph{arXiv preprint arXiv:2106.13112}, 2021.

\bibitem{devlin2019bert}
J.~Devlin, M.-W. Chang, K.~Lee, and K.~Toutanova, ``Bert: Pre-training of deep bidirectional transformers for language understanding,'' in \emph{NAACL}, 2019.

\bibitem{strubell2019energy}
E.~Strubell, A.~Ganesh, and A.~McCallum, ``Energy and policy considerations for deep learning in nlp,'' in \emph{ACL}, 2019.

\bibitem{Sun2017RevisitingUE}
C.~Sun, A.~Shrivastava, S.~Singh, and A.~Gupta, ``Revisiting unreasonable effectiveness of data in deep learning era,'' in \emph{ICCV}, 2017.

\bibitem{Patterson2021CarbonEA}
D.~Patterson, J.~Gonzalez, Q.~V. Le, C.~Liang, L.-M. Mungu{\'i}a, D.~Rothchild, D.~R. So, M.~Texier, and J.~Dean, ``Carbon emissions and large neural network training,'' \emph{arXiv preprint arXiv:2104.10350}, 2021.

\bibitem{frankle2019lottery}
J.~Frankle and M.~Carbin, ``The lottery ticket hypothesis: Finding sparse, trainable neural networks,'' in \emph{ICLR}, 2019.

\bibitem{Gong2019EfficientTO}
L.~Gong, D.~He, Z.~Li, T.~Qin, L.~Wang, and T.-Y. Liu, ``Efficient training of bert by progressively stacking,'' in \emph{ICML}, 2019.

\bibitem{Tan2021EfficientNetV2SM}
M.~Tan and Q.~V. Le, ``Efficientnetv2: Smaller models and faster training,'' in \emph{ICML}, 2021.

\bibitem{rombach2022highresolution}
R.~Rombach, A.~Blattmann, D.~Lorenz, P.~Esser, and B.~Ommer, ``High-resolution image synthesis with latent diffusion models,'' in \emph{CVPR}, 2022, pp. 10\,684--10\,695.

\bibitem{Fahlman1989TheCL}
S.~E. Fahlman and C.~Lebiere, ``The cascade-correlation learning architecture,'' in \emph{NeurIPS}, 1989.

\bibitem{Lengell1996TrainingML}
R.~Lengell{\'e} and T.~Denoeux, ``Training mlps layer by layer using an objective function for internal representations,'' \emph{Neural Networks}, vol.~9, 1996.

\bibitem{Hinton2006AFL}
G.~E. Hinton, S.~Osindero, and Y.~W. Teh, ``A fast learning algorithm for deep belief nets,'' \emph{Neural Computation}, vol.~18, 2006.

\bibitem{Bengio2006GreedyLT}
Y.~Bengio, P.~Lamblin, D.~Popovici, and H.~Larochelle, ``Greedy layer-wise training of deep networks,'' in \emph{NeurIPS}, 2006.

\bibitem{Simonyan2015VeryDC}
K.~Simonyan and A.~Zisserman, ``Very deep convolutional networks for large-scale image recognition,'' in \emph{ICLR}, 2014.

\bibitem{Smith2016GradualDO}
L.~N. Smith, E.~M. Hand, and T.~Doster, ``Gradual dropin of layers to train very deep neural networks,'' in \emph{CVPR}, 2016.

\bibitem{Karras2018ProgressiveGO}
T.~Karras, T.~Aila, S.~Laine, and J.~Lehtinen, ``Progressive growing of gans for improved quality, stability, and variation,'' in \emph{ICLR}, 2018.

\bibitem{Wang2017DeepGL}
G.~Wang, X.~Xie, J.~Lai, and J.~Zhuo, ``Deep growing learning,'' in \emph{ICCV}, 2017, pp. 2812--2820.

\bibitem{Chen2016Net2NetAL}
T.~Chen, I.~Goodfellow, and J.~Shlens, ``Net2net: Accelerating learning via knowledge transfer,'' in \emph{ICLR}, 2016.

\bibitem{Wei2016NetworkM}
T.~Wei, C.~Wang, Y.~Rui, and C.~W. Chen, ``Network morphism,'' in \emph{ICML}, 2016.

\bibitem{Wei2021ModularizedMO}
T.~Wei, C.~Wang, and C.~W. Chen, ``Modularized morphing of deep convolutional neural networks: A graph approach,'' \emph{IEEE Transactions on Computers}, 2021.

\bibitem{Li2020ShallowtoDeepTF}
B.~Li, Z.~Wang, H.~Liu, Y.~Jiang, Q.~Du, T.~Xiao, H.~Wang, and J.~Zhu, ``Shallow-to-deep training for neural machine translation,'' in \emph{EMNLP}, 2020.

\bibitem{Yang2020ProgressivelyS2}
C.~Yang, S.~Wang, C.~Yang, Y.~Li, R.~He, and J.~Zhang, ``Progressively stacking 2.0: A multi-stage layerwise training method for bert training speedup,'' \emph{arXiv preprint arXiv:2011.13635}, 2020.

\bibitem{zhang2020accelerating}
M.~Zhang and Y.~He, ``Accelerating training of transformer-based language models with progressive layer dropping,'' in \emph{NeurIPS}, 2020.

\bibitem{Gu2021OnTT}
X.~Gu, L.~Liu, H.~Yu, J.~Li, C.~Chen, and J.~Han, ``On the transformer growth for progressive bert training,'' in \emph{NAACL}, 2021.

\bibitem{You2020L2GCNLA}
Y.~You, T.~Chen, Z.~Wang, and Y.~Shen, ``L2-gcn: Layer-wise and learned efficient training of graph convolutional networks,'' in \emph{CVPR}, 2020.

\bibitem{Wang2021StackRecET}
J.~Wang, F.~Yuan, J.~Chen, Q.~Wu, M.~Yang, Y.~Sun, and G.~Zhang, ``Stackrec: Efficient training of very deep sequential recommender models by iterative stacking,'' in \emph{ACM SIGIR}, 2021.

\bibitem{zoph2016neural}
B.~Zoph and Q.~V. Le, ``Neural architecture search with reinforcement learning,'' in \emph{{ICLR}}, 2017.

\bibitem{baker2016designing}
B.~Baker, O.~Gupta, N.~Naik, and R.~Raskar, ``Designing neural network architectures using reinforcement learning,'' in \emph{{ICLR}}, 2017.

\bibitem{Tan2018MnasNetPN}
M.~Tan, B.~Chen, R.~Pang, V.~Vasudevan, M.~Sandler, A.~Howard, and Q.~V. Le, ``Mnasnet: Platform-aware neural architecture search for mobile,'' in \emph{{CVPR}}, 2019.

\bibitem{liu2018progressive}
C.~Liu, B.~Zoph, M.~Neumann, J.~Shlens, W.~Hua, L.-J. Li, L.~Fei-Fei, A.~Yuille, J.~Huang, and K.~Murphy, ``Progressive neural architecture search,'' in \emph{{ECCV}}, 2018.

\bibitem{Bergstra2011AlgorithmsFH}
J.~Bergstra, R.~Bardenet, Y.~Bengio, and B.~K{\'e}gl, ``Algorithms for hyper-parameter optimization,'' in \emph{NeurIPS}, 2011.

\bibitem{Bergstra2012RandomSF}
J.~Bergstra and Y.~Bengio, ``Random search for hyper-parameter optimization,'' \emph{JMLR}, vol.~13, 2012.

\bibitem{Cubuk2019AutoAugmentLA}
E.~D. Cubuk, B.~Zoph, D.~Man{\'e}, V.~Vasudevan, and Q.~V. Le, ``Autoaugment: Learning augmentation strategies from data,'' in \emph{CVPR}, 2019.

\bibitem{tang2022learning}
T.~Tang, C.~Li, G.~Wang, K.~Yu, X.~Chang, and X.~Liang, ``Learning self-regularized adversarial views for self-supervised vision transformers,'' \emph{arXiv preprint arXiv:2210.08458}, 2022.

\bibitem{Wu2018LearningTT}
L.~Wu, F.~Tian, Y.~Xia, Y.~Fan, T.~Qin, J.~Lai, and T.-Y. Liu, ``Learning to teach with dynamic loss functions,'' in \emph{NeurIPS}, 2018.

\bibitem{Xu2019AutoLossLD}
H.~Xu, H.~Zhang, Z.~Hu, X.~Liang, R.~Salakhutdinov, and E.~P. Xing, ``Autoloss: Learning discrete schedules for alternate optimization,'' in \emph{ICLR}, 2019.

\bibitem{li2020autosegloss}
H.~Li, C.~Tao, X.~Zhu, X.~Wang, G.~Huang, and J.~Dai, ``Auto seg-loss: Searching metric surrogates for semantic segmentation,'' in \emph{ICLR}, 2021.

\bibitem{Liu2018DARTSDA}
H.~Liu, K.~Simonyan, and Y.~Yang, ``{DARTS:} differentiable architecture search,'' in \emph{{ICLR}}, 2019.

\bibitem{Cai2018ProxylessNASDN}
H.~Cai, L.~Zhu, and S.~Han, ``Proxylessnas: Direct neural architecture search on target task and hardware,'' in \emph{{ICLR}}, 2019.

\bibitem{brock2017smash}
A.~Brock, T.~Lim, J.~M. Ritchie, and N.~Weston, ``{SMASH:} one-shot model architecture search through hypernetworks,'' in \emph{{ICLR}}, 2018.

\bibitem{pham2018enas}
H.~Pham, M.~Guan, B.~Zoph, Q.~Le, and J.~Dean, ``Efficient neural architecture search via parameters sharing,'' in \emph{{ICML}}, 2018.

\bibitem{guo2020single}
Z.~Guo, X.~Zhang, H.~Mu, W.~Heng, Z.~Liu, Y.~Wei, and J.~Sun, ``Single path one-shot neural architecture search with uniform sampling,'' in \emph{ECCV}, 2020.

\bibitem{li2019blockwisely}
C.~Li, J.~Peng, L.~Yuan, G.~Wang, X.~Liang, L.~Lin, and X.~Chang, ``Block-wisely supervised neural architecture search with knowledge distillation,'' in \emph{{CVPR}}, 2020.

\bibitem{peng2021pi}
J.~Peng, J.~Zhang, C.~Li, G.~Wang, X.~Liang, and L.~Lin, ``Pi-nas: Improving neural architecture search by reducing supernet training consistency shift,'' in \emph{ICCV}, 2021.

\bibitem{wang2023dna}
G.~Wang, C.~Li, L.~Yuan, J.~Peng, X.~Xian, X.~Liang, X.~Chang, and L.~Lin, ``Dna family: Boosting weight-sharing nas with block-wise supervisions,'' \emph{IEEE Transactions on Pattern Analysis and Machine Intelligence}, 2023.

\bibitem{Li2020DADADA}
Y.~Li, G.~Hu, Y.~Wang, T.~M. Hospedales, N.~M. Robertson, and Y.~Yang, ``Dada: Differentiable automatic data augmentation,'' in \emph{ECCV}, 2020.

\bibitem{Wen2020AutoGrowAL}
W.~Wen, F.~Yan, and H.~H. Li, ``Autogrow: Automatic layer growing in deep convolutional networks,'' in \emph{KDD}, 2020.

\bibitem{Dong2020TowardsAR}
C.~Dong, L.~Liu, Z.~Li, and J.~Shang, ``Towards adaptive residual network training: A neural-ode perspective,'' in \emph{ICML}, 2020.

\bibitem{Kim2021TheLC}
H.~Kim, G.~Papamakarios, and A.~Mnih, ``The lipschitz constant of self-attention,'' in \emph{ICML}, 2021.

\bibitem{li2022automated}
C.~Li, B.~Zhuang, G.~Wang, X.~Liang, X.~Chang, and Y.~Yang, ``Automated progressive learning for efficient training of vision transformers,'' in \emph{Proceedings of the IEEE/CVF Conference on Computer Vision and Pattern Recognition}, 2022, pp. 12\,486--12\,496.

\bibitem{larsson2016fractalnet}
G.~Larsson, M.~Maire, and G.~Shakhnarovich, ``Fractalnet: Ultra-deep neural networks without residuals,'' in \emph{{ICLR}}, 2017.

\bibitem{Huang2018MultiScaleDN}
G.~Huang, D.~Chen, T.~Li, F.~Wu, L.~van~der Maaten, and K.~Q. Weinberger, ``Multi-scale dense networks for resource efficient image classification,'' in \emph{{ICLR}}, 2018.

\bibitem{hu2019learning}
H.~Hu, D.~Dey, M.~Hebert, and J.~A. Bagnell, ``Learning anytime predictions in neural networks via adaptive loss balancing,'' in \emph{{AAAI}}, 2019.

\bibitem{Lee2018AnytimeNP}
H.~Lee and J.~Shin, ``Anytime neural prediction via slicing networks vertically,'' \emph{arXiv preprint arXiv:1807.02609}, 2018.

\bibitem{yu2019slimmable}
J.~Yu, L.~Yang, N.~Xu, J.~Yang, and T.~Huang, ``Slimmable neural networks,'' in \emph{ICLR}, 2019.

\bibitem{Yu2019UniversallySN}
J.~Yu and T.~S. Huang, ``Universally slimmable networks and improved training techniques,'' in \emph{{ICCV}}, 2019.

\bibitem{yu2019autoslim}
J.~Yu and T.~Huang, ``Autoslim: Towards one-shot architecture search for channel numbers,'' in \emph{NeurIPS workshop}, 2019.

\bibitem{Cai2020Once_for_All}
H.~Cai, C.~Gan, T.~Wang, Z.~Zhang, and S.~Han, ``Once-for-all: Train one network and specialize it for efficient deployment,'' in \emph{ICLR}, 2020.

\bibitem{Yu2020BigNASSU}
J.~Yu, P.~Jin, H.~Liu, G.~Bender, P.-J. Kindermans, M.~Tan, T.~Huang, X.~Song, and Q.~V. Le, ``Bignas: Scaling up neural architecture search with big single-stage models,'' in \emph{{ECCV}}, 2020.

\bibitem{chen2021autoformer}
M.~Chen, H.~Peng, J.~Fu, and H.~Ling, ``Autoformer: Searching transformers for visual recognition,'' in \emph{ICCV}, 2021.

\bibitem{Li2019ImprovedTF}
H.~Li, H.~Zhang, X.~Qi, R.~Yang, and G.~Huang, ``Improved techniques for training adaptive deep networks,'' in \emph{{ICCV}}, 2019.

\bibitem{li2021dynamic}
C.~Li, G.~Wang, B.~Wang, X.~Liang, Z.~Li, and X.~Chang, ``Dynamic slimmable network,'' in \emph{CVPR}, 2021.

\bibitem{wang2021not}
Y.~Wang, R.~Huang, S.~Song, Z.~Huang, and G.~Huang, ``Not all images are worth 16x16 words: Dynamic vision transformers with adaptive sequence length,'' in \emph{NeurIPS}, 2021.

\bibitem{li2021ds}
C.~Li, G.~Wang, B.~Wang, X.~Liang, Z.~Li, and X.~Chang, ``Ds-net++: Dynamic weight slicing for efficient inference in cnns and vision transformers,'' \emph{IEEE Transactions on Pattern Analysis and Machine Intelligence}, vol.~45, no.~4, pp. 4430--4446, 2022.

\bibitem{jiang2023dynamic}
Z.~Jiang, C.~Li, X.~Chang, L.~Chen, J.~Zhu, and Y.~Yang, ``Dynamic slimmable denoising network,'' \emph{IEEE Transactions on Image Processing}, vol.~32, pp. 1583--1598, 2023.

\bibitem{hou2020dynabert}
L.~Hou, L.~Shang, X.~Jiang, and Q.~Liu, ``Dynabert: Dynamic bert with adaptive width and depth,'' in \emph{NeurIPS}, 2020.

\bibitem{ho2020denoising}
J.~Ho, A.~Jain, and P.~Abbeel, ``Denoising diffusion probabilistic models,'' \emph{arXiv preprint arXiv:2006.11239}, 2020.

\bibitem{song2021scorebased}
Y.~Song, J.~Sohl-Dickstein, D.~P. Kingma, A.~Kumar, S.~Ermon, and B.~Poole, ``Score-based generative modeling through stochastic differential equations,'' \emph{arXiv preprint arXiv:2011.13456}, 2021.

\bibitem{radford2021learning}
A.~Radford, J.~W. Kim, C.~Hallacy, A.~Ramesh, G.~Goh, S.~Agarwal, G.~Sastry, A.~Askell, P.~Mishkin, J.~Clark \emph{et~al.}, ``Learning transferable visual models from natural language supervision,'' in \emph{ICML}, 2021.

\bibitem{ramesh2021zero}
A.~Ramesh, M.~Pavlov, G.~Goh, S.~Gray, C.~Voss, A.~Radford, M.~Chen, and I.~Sutskever, ``Zero-shot text-to-image generation,'' \emph{arXiv preprint arXiv:2102.12092}, 2021.

\bibitem{saharia2022photorealistic}
C.~Saharia, W.~Chan, S.~Saxena, L.~Li, J.~Whang, E.~Denton, S.~K.~S. Ghasemipour, B.~K. Ayan, T.~Salimans, J.~Ho, D.~J. Fleet \emph{et~al.}, ``Photorealistic text-to-image diffusion models with deep language understanding,'' \emph{arXiv preprint arXiv:2205.11487}, 2022.

\bibitem{ramesh2022hierarchical}
A.~Ramesh, P.~Dhariwal, A.~Nichol, C.~Chu, and M.~Chen, ``Hierarchical text-conditional image generation with clip latents,'' in \emph{ICML}, 2022.

\bibitem{zaken2021bitfit}
E.~B. Zaken, S.~Ravfogel, and Y.~Goldberg, ``Bitfit: Simple parameter-efficient fine-tuning for transformer-based masked language-models,'' \emph{arXiv preprint arXiv:2106.10199}, 2021.

\bibitem{hu2021lora}
E.~J. Hu, Y.~Shen, P.~Wallis, Z.~Allen-Zhu, Y.~Li, S.~Wang, L.~Wang, and W.~Chen, ``Lora: Low-rank adaptation of large language models,'' \emph{arXiv preprint arXiv:2106.09685}, 2021.

\bibitem{li2021prefix}
X.~L. Li and P.~Liang, ``Prefix-tuning: Optimizing continuous prompts for generation,'' \emph{arXiv preprint arXiv:2101.00190}, 2021.

\bibitem{Dollr2021FastAA}
P.~Doll{\'a}r, M.~Singh, and R.~B. Girshick, ``Fast and accurate model scaling,'' in \emph{CVPR}, 2021.

\bibitem{chang2018multi}
B.~Chang, L.~Meng, E.~Haber, F.~Tung, and D.~Begert, ``Multi-level residual networks from dynamical systems view,'' in \emph{ICLR}, 2018.

\bibitem{BYOL}
J.-B. Grill, F.~Strub, F.~Altché, C.~Tallec, P.~H. Richemond, E.~Buchatskaya, C.~Doersch, B.~A. Pires, Z.~D. Guo, M.~G. Azar, B.~Piot, K.~Kavukcuoglu, R.~Munos, and M.~Valko, ``Bootstrap your own latent: A new approach to self-supervised learning,'' in \emph{NeurIPS}, 2020.

\bibitem{guo2020bootstrap}
D.~Guo, B.~A. Pires, B.~Piot, J.-b. Grill, F.~Altch{\'e}, R.~Munos, and M.~G. Azar, ``Bootstrap latent-predictive representations for multitask reinforcement learning,'' in \emph{{ICML}}, 2020.

\bibitem{He2020MomentumCF}
K.~He, H.~Fan, Y.~Wu, S.~Xie, and R.~B. Girshick, ``Momentum contrast for unsupervised visual representation learning,'' in \emph{CVPR}, 2020.

\bibitem{laine2016temporal}
S.~Laine and T.~Aila, ``Temporal ensembling for semi-supervised learning,'' \emph{arXiv preprint arXiv:1610.02242}, 2016.

\bibitem{tarvainen2017mean}
A.~Tarvainen and H.~Valpola, ``Mean teachers are better role models: Weight-averaged consistency targets improve semi-supervised deep learning results,'' in \emph{{NeurIPS}}, 2017.

\bibitem{deb2014multi}
K.~Deb, ``Multi-objective optimization,'' in \emph{Search methodologies}.\hskip 1em plus 0.5em minus 0.4em\relax Springer, 2014.

\bibitem{Tan2019EfficientNetRM}
M.~Tan and Q.~V. Le, ``Efficientnet: Rethinking model scaling for convolutional neural networks,'' in \emph{ICML}, 2019.

\bibitem{Li2021BossNASEH}
C.~Li, T.~Tang, G.~Wang, J.~Peng, B.~Wang, X.~Liang, and X.~Chang, ``Bossnas: Exploring hybrid cnn-transformers with block-wisely self-supervised neural architecture search,'' in \emph{ICCV}, 2021.

\bibitem{Srivastava2014DropoutAS}
N.~Srivastava, G.~E. Hinton, A.~Krizhevsky, I.~Sutskever, and R.~Salakhutdinov, ``Dropout: a simple way to prevent neural networks from overfitting,'' \emph{J. Mach. Learn. Res.}, vol.~15, 2014.

\bibitem{Huang2016DeepNW}
G.~Huang, Y.~Sun, Z.~Liu, D.~Sedra, and K.~Q. Weinberger, ``Deep networks with stochastic depth,'' in \emph{ECCV}, 2016.

\bibitem{Cai2021NetworkAF}
H.~Cai, C.~Gan, J.~Lin, and S.~Han, ``Network augmentation for tiny deep learning,'' \emph{arXiv preprint arXiv:2110.08890}, 2021.

\bibitem{ruiz2023dreambooth}
N.~Ruiz, Y.~Li, V.~Jampani, Y.~Pritch, M.~Rubinstein, and K.~Aberman, ``Dreambooth: Fine tuning text-to-image diffusion models for subject-driven generation,'' in \emph{CVPR}, 2023, pp. 22\,500--22\,510.

\bibitem{evci2020rigging}
U.~Evci, T.~Gale, J.~Menick, P.~S. Castro, and E.~Elsen, ``Rigging the lottery: Making all tickets winners,'' in \emph{ICML}, 2020.

\bibitem{tanaka2020pruning}
H.~Tanaka, D.~Kunin, D.~L. Yamins, and S.~Ganguli, ``Pruning neural networks without any data by iteratively conserving synaptic flow,'' in \emph{NeurIPS}, vol.~33, 2020.

\bibitem{chen2021neural}
W.~Chen, X.~Gong, and Z.~Wang, ``Neural architecture search on imagenet in four gpu hours: A theoretically inspired perspective,'' in \emph{ICLR}, 2021.

\bibitem{xiao2020disentangling}
L.~Xiao, J.~Pennington, and S.~Schoenholz, ``Disentangling trainability and generalization in deep neural networks,'' in \emph{ICML}.\hskip 1em plus 0.5em minus 0.4em\relax PMLR, 2020, pp. 10\,462--10\,472.

\bibitem{jacot2018neural}
A.~Jacot, F.~Gabriel, and C.~Hongler, ``Neural tangent kernel: Convergence and generalization in neural networks,'' in \emph{NeurIPS}, vol.~31, 2018.

\bibitem{lee2019wide}
J.~Lee, L.~Xiao, S.~Schoenholz, Y.~Bahri, R.~Novak, J.~Sohl-Dickstein, and J.~Pennington, ``Wide neural networks of any depth evolve as linear models under gradient descent,'' in \emph{NeurIPS}, vol.~32, 2019.

\bibitem{li2023zico}
G.~Li, Y.~Yang, K.~Bhardwaj, and R.~Marculescu, ``Zico: Zero-shot nas via inverse coefficient of variation on gradients,'' \emph{arXiv preprint arXiv:2301.11300}, 2023.

\bibitem{lewkowycz2020large}
A.~Lewkowycz, Y.~Bahri, E.~Dyer, J.~Sohl-Dickstein, and G.~Gur-Ari, ``The large learning rate phase of deep learning: the catapult mechanism,'' \emph{arXiv preprint arXiv:2003.02218}, 2020.

\bibitem{deng2009imagenet}
J.~Deng, W.~Dong, R.~Socher, L.-J. Li, K.~Li, and L.~Fei-Fei, ``Imagenet: A large-scale hierarchical image database,'' in \emph{{CVPR}}, 2009.

\bibitem{Krizhevsky09cifar}
A.~Krizhevsky and G.~Hinton, ``Learning multiple layers of features from tiny images,'' \emph{Master's thesis, Department of Computer Science, University of Toronto}, 2009.

\bibitem{micikevicius2018mixed}
P.~Micikevicius, S.~Narang, J.~Alben, G.~Diamos, E.~Elsen, D.~Garcia, B.~Ginsburg, M.~Houston, O.~Kuchaiev, G.~Venkatesh \emph{et~al.}, ``Mixed precision training,'' in \emph{ICLR}, 2018.

\bibitem{liao2022artbench}
P.~Liao, X.~Li, X.~Liu, and K.~Keutzer, ``The artbench dataset: Benchmarking generative models with artworks,'' \emph{arXiv preprint arXiv:2206.11404}, 2022.

\bibitem{wah2011caltech}
C.~Wah, S.~Branson, P.~Welinder, P.~Perona, and S.~Belongie, ``The caltech-ucsd birds-200-2011 dataset,'' 2011.

\bibitem{nilsback2008automated}
M.-E. Nilsback and A.~Zisserman, ``Automated flower classification over a large number of classes,'' in \emph{2008 Sixth Indian conference on computer vision, graphics \& image processing}.\hskip 1em plus 0.5em minus 0.4em\relax IEEE, 2008, pp. 722--729.

\bibitem{krause20133d}
J.~Krause, M.~Stark, J.~Deng, and L.~Fei-Fei, ``3d object representations for fine-grained categorization,'' in \emph{ICCV Workshops}, 2013, pp. 554--561.

\bibitem{chen2022adaptformer}
S.~Chen, C.~Ge, Z.~Tong, J.~Wang, Y.~Song, J.~Wang, and P.~Luo, ``Adaptformer: Adapting vision transformers for scalable visual recognition,'' in \emph{NeurIPS}, vol.~35, 2022, pp. 16\,664--16\,678.

\bibitem{jia2022visual}
M.~Jia, L.~Tang, B.-C. Chen, C.~Cardie, S.~Belongie, B.~Hariharan, and S.-N. Lim, ``Visual prompt tuning,'' in \emph{ECCV}.\hskip 1em plus 0.5em minus 0.4em\relax Springer, 2022, pp. 709--727.

\bibitem{caron2021emerging}
M.~Caron, H.~Touvron, I.~Misra, H.~J{\'e}gou, J.~Mairal, P.~Bojanowski, and A.~Joulin, ``Emerging properties in self-supervised vision transformers,'' in \emph{ICCV}, 2021, pp. 9650--9660.

\bibitem{Bachlechner2020ReZeroIA}
T.~C. Bachlechner, B.~P. Majumder, H.~H. Mao, G.~Cottrell, and J.~McAuley, ``Rezero is all you need: Fast convergence at large depth,'' \emph{arXiv preprint arXiv:2003.04887}, 2020.

\bibitem{bossard2014food}
L.~Bossard, M.~Guillaumin, and L.~Van~Gool, ``Food-101--mining discriminative components with random forests,'' in \emph{ECCV}.\hskip 1em plus 0.5em minus 0.4em\relax Springer, 2014, pp. 446--461.

\bibitem{loshchilov2018decoupled}
I.~Loshchilov and F.~Hutter, ``Decoupled weight decay regularization,'' in \emph{ICLR}, 2019.

\bibitem{Cubuk2020RandaugmentPA}
E.~D. Cubuk, B.~Zoph, J.~Shlens, and Q.~V. Le, ``Randaugment: Practical automated data augmentation with a reduced search space,'' in \emph{CVPR Workshop}, 2020.

\bibitem{Zhang2018mixupBE}
H.~Zhang, M.~Ciss{\'e}, Y.~Dauphin, and D.~Lopez-Paz, ``mixup: Beyond empirical risk minimization,'' in \emph{ICLR}, 2018.

\bibitem{Yun2019CutMixRS}
S.~Yun, D.~Han, S.~J. Oh, S.~Chun, J.~Choe, and Y.~J. Yoo, ``Cutmix: Regularization strategy to train strong classifiers with localizable features,'' in \emph{ICCV}, 2019.

\bibitem{Zhong2020RandomED}
Z.~Zhong, L.~Zheng, G.~Kang, S.~Li, and Y.~Yang, ``Random erasing data augmentation,'' in \emph{AAAI}, 2020.

\bibitem{Hoffer2020AugmentYB}
E.~Hoffer, T.~Ben-Nun, I.~Hubara, N.~Giladi, T.~Hoefler, and D.~Soudry, ``Augment your batch: Improving generalization through instance repetition,'' in \emph{CVPR}, 2020.

\bibitem{jiang2021all}
Z.~Jiang, Q.~Hou, L.~Yuan, D.~Zhou, Y.~Shi, X.~Jin, A.~Wang, and J.~Feng, ``All tokens matter: Token labeling for training better vision transformers,'' \emph{arXiv preprint arXiv:2104.10858}, 2021.

\end{thebibliography}

\newpage
\appendices

\section{Definition of Compared Growth Operators}
Given a smaller network $\bm\psi_s$ and a larger network $\bm\psi_\ell$, a growth operator $\bm\zeta$ maps the parameters of the smaller one $\bm\omega_s$ to the parameters of the larger one $\bm\omega_\ell$ by: $\bm\omega_\ell = \bm\zeta(\bm\omega_s)$. Let $\bm\omega_\ell^i$ denotes the parameters of the $i$-th layer in $\bm\psi_\ell$\footnote{In our default setting, $i$ begins from the layer near the classifier.}. We consider several $\bm\zeta$ in depth dimension that maps $\bm\omega_s$ to layer $i$ of $\bm\psi_\ell$ by: $\bm\omega_\ell^i = \bm\zeta(\bm\omega_s, i)$. 

\noindent\textbf{RandInit.} \textit{RandInit} copies the original layers in $\bm\psi_s$ and random initialize the newly added layers:
\begin{equation}
    \bm\zeta_\textit{RandInit}(\bm\omega_s, i) = \left\{
    \begin{aligned}
    \bm\omega_s^i, ~~~~~~~~~~~& i\leq l_s\\
    \textit{RandInit}, ~~& i > l_s.\\
    \end{aligned}
    \right.
\end{equation}

\noindent\textbf{Stacking.} \textit{Stacking} duplicates the original layers and directly stacks the duplicated ones on top of them:
\begin{equation}
    \bm\zeta_\textit{Stacking}(\bm\omega_s, i) = 
    \bm\omega_s^{i\bmod{l_s}}.
\end{equation}

\noindent\textbf{Interpolation.} \textit{Interpolation} interpolates new layers of $\bm\psi_\ell$ in between original ones and copy the weights from their nearest neighbor in $\bm\psi_s$:
\begin{equation}
    \bm\zeta_\textit{Interpolation}(\bm\omega_s, i) = \bm\omega_s^{\lfloor i/l_s\rfloor}.
\end{equation}

\section{Implementation Details}

Our ImageNet training settings of AutoProg-\textit{One} follow closely to the original training settings of DeiT~\cite{touvron2020deit} and VOLO~\cite{Yuan2021VOLOVO}, respectively. We use the AdamW optimizer~\cite{loshchilov2018decoupled} with an initial learning rate of 1e-3, a total batch size of 1024 and a weight decay rate of 5e-2 for both architectures. The learning rate decays following a cosine schedule with 20 epochs warm-up for VOLO models and 5 epochs warm-up for DeiT models. For both architectures, we use exponential moving average with best momentum factor in $\{0.998, 0.9986, 0.999, 0.9996\}$.

For DeiT training, we use RandAugment~\cite{Cubuk2020RandaugmentPA} with 9 magnitude and 0.5 magnitude std., mixup~\cite{Zhang2018mixupBE} with 0.8 probability, cutmix~\cite{Yun2019CutMixRS} with 1.0 probability, random erasing~\cite{Zhong2020RandomED} with 0.25 probability, stochastic depth~\cite{Huang2016DeepNW} with 0.1 probability and repeated augmentation~\cite{Hoffer2020AugmentYB}.

For VOLO training, we use RandAugment~\cite{Cubuk2020RandaugmentPA}, random erasing~\cite{Zhong2020RandomED}, stochastic depth~\cite{Huang2016DeepNW}, token labeling with MixToken~\cite{jiang2021all}, with magnitude of RandAugment, probability of random erasing and stochastic depth adjusted by Adaptive Regularization.

\noindent\textbf{Adaptive Regularization.}

The detailed settings of Adaptive Regularization for VOLO progressive training is shown in \cref{tab:AdaReg}. These hyper-parameters are set heuristically regarding the model size. They perform fairly well in our experiments, but could still be sub-optimal.
\begin{table}[ht]
    \centering
    \footnotesize
    \setlength{\tabcolsep}{10pt}
    \begin{tabular}{l|cc|cc}
    \toprule
         \multirow{2}{*}{Regularization}&  \multicolumn{2}{c|}{D0} &   \multicolumn{2}{c}{D1}\\
         & min & max & min & max\\
         \midrule
         RandAugment~\cite{Cubuk2020RandaugmentPA} & 4.5 & 9 & 4.5 & 9\\
         Random Erasing~\cite{Zhong2020RandomED}& 0 & 0.25 & 0.0625 & 0.25\\
         Stoch. Depth~\cite{Huang2016DeepNW} & 0 & 0.1 & 0.1 & 0.2\\
    \bottomrule
    \end{tabular}\vspace{-5pt}
    \caption{Adaptive Regularization Settings (magnitude of RandAugment~\cite{Cubuk2020RandaugmentPA}, probability of Random Erasing~\cite{Zhong2020RandomED} and Stochastic Depth~\cite{Huang2016DeepNW}) for progressive training of VOLO models.}
    \label{tab:AdaReg}
\end{table}

\begin{table*}[t]
    \footnotesize
    \centering
    \setlength{\tabcolsep}{7pt}
        \begin{tabular}{l|l|cc|cc|c|c}
        \toprule
        Model  & \makecell[l]{Training\\scheme}  &\makecell{FLOPs\\(avg. per step)} & Speedup & \makecell{Runtime\\(GPU Hours)} & Speedup    & \makecell{Top-1\\(\%)} & \makecell{Top-1@288\\(\%)}\\
        \midrule
        \multicolumn{4}{l}{\textbf{\textit{100 epochs}}}\\
        \midrule
        \multirow{3}{*}{DeiT-S~\cite{touvron2020deit}}          & Original     &   4.6G & \na &71& \na           & 74.1      & 74.6\\ %
        & Prog       &2.4G&+91.6\%&                            46
        & +53.6\% & 72.6      &  73.2\\ %
        & \cellcolor{Light}AutoProg-\textit{One}                                             &\cellcolor{Light}2.8G&\cellcolor{Light}+62.0\%& \cellcolor{Light} 50  &\cellcolor{Light} +40.7\%                   & \cellcolor{Light}\textbf{74.4}      &\cellcolor{Light} \textbf{74.9}\\ %
        \arrayrulecolor{lightgray}\hline\arrayrulecolor{black}
        \multirow{4}{*}{VOLO-D1~\cite{Yuan2021VOLOVO}}          & Original     &6.8G&\na&150& \na           & 82.6      &83.0\\ %
        & Prog                                                     &3.7G&+84.7\%&93& +60.9\% & 81.7      &82.1\\ %
        & \cellcolor{Light}AutoProg-\textit{One}                                     &\cellcolor{Light}3.3G&\cellcolor{Light}+104.2\%&\cellcolor{Light}91&\cellcolor{Light} +65.6\%                   &\cellcolor{Light} \textbf{82.8}      &\cellcolor{Light}\textbf{83.2}\\ %
        & \cellcolor{Light}AutoProg-\textit{One} 0.4$\Omega$                                     &\cellcolor{Light}2.9G&\cellcolor{Light}\textbf{+132.2\%}&\cellcolor{Light}81&\cellcolor{Light} \textbf{+85.1\%}          & \cellcolor{Light}82.7      & \cellcolor{Light}83.1 \\ %
        \arrayrulecolor{lightgray}\hline\arrayrulecolor{black}
        \multirow{3}{*}{VOLO-D2~\cite{Yuan2021VOLOVO}}          & Original     &14.1G&\na&277& \na                       & 83.6      & 84.1\\ %
        & Prog                                                     &7.5G&+87.7\%&180& +54.4\% & 82.9      & 83.3\\ %
        & \cellcolor{Light}AutoProg-\textit{One}                                                 &\cellcolor{Light}8.3G&\cellcolor{Light}+68.7\%&\cellcolor{Light}191&\cellcolor{Light} +45.3\%                   &\cellcolor{Light} \textbf{83.8}      &\cellcolor{Light} \textbf{84.2}\\ %
        \midrule
        \multicolumn{4}{l}{\textbf{\textit{300 epochs}}}\\
        \midrule
        \multirow{2}{*}{DeiT-Tiny~\cite{touvron2020deit}}       &  Original    &1.2G&\na&144& \na           & 72.2      & 72.9\\ %
        & \cellcolor{Light}AutoProg-\textit{One}                                                 &\cellcolor{Light}0.7G&\cellcolor{Light}+82.1\%&\cellcolor{Light}95& \cellcolor{Light}   +51.2\%                       & \cellcolor{Light} \textbf{72.4}  & \cellcolor{Light}\textbf{73.0}  \\ %
        \arrayrulecolor{lightgray}\hline\arrayrulecolor{black}
        \multirow{2}{*}{DeiT-S~\cite{touvron2020deit}}          &  Original    &4.6G&\na&213& \na           & 79.8      &  80.1\\ %
        & \cellcolor{Light}AutoProg-\textit{One}                                                 &\cellcolor{Light}2.8G&\cellcolor{Light}+62.0\%&\cellcolor{Light}150&\cellcolor{Light} +42.0\%                   &\cellcolor{Light} 79.8      & \cellcolor{Light} 80.1 \\ %
        \arrayrulecolor{lightgray}\hline\arrayrulecolor{black}
        \multirow{2}{*}{VOLO-D1~\cite{Yuan2021VOLOVO}}          & Original     &6.8G&\na&487& \na           & 84.2      & 84.4 \\ %
        &\cellcolor{Light}AutoProg-\textit{One}                                                 &\cellcolor{Light}4.0G&\cellcolor{Light}+68.9\%&\cellcolor{Light}327&\cellcolor{Light} +48.9\%                   &\cellcolor{Light} \textbf{84.3}      &\cellcolor{Light}\textbf{84.6}\\ %
        \arrayrulecolor{lightgray}\hline\arrayrulecolor{black}
        \multirow{2}{*}{VOLO-D2~\cite{Yuan2021VOLOVO}}          &Original      &14.1G&\na&863& \na           & 85.2      & 85.1\\
        &\cellcolor{Light}AutoProg-\textit{One}                                                 &\cellcolor{Light}8.8G&\cellcolor{Light}+60.7\%&\cellcolor{Light}605&\cellcolor{Light} +42.7\%                   &\cellcolor{Light} 85.2     & \cellcolor{Light} \textbf{85.2}\\
        \bottomrule
    \end{tabular}%
    \caption{Detailed results of efficient training on ImageNet. Best results are marked with \textbf{Bold}; our method or default settings are highlighted in \colorbox{Light}{purple}. Top-1@288 denotes Top-1 Accuracy when directly testing on 288$\times$288 input size, \textit{without} finetuning. Runtime is rounded to integer.}
    \label{tab:appimagenet}
\end{table*}

\noindent\textbf{Growth Space $\bm\Lambda_{\bm{k}}$ in Each Stage.}
We find emprically that the elastic supernet converges faster when the number of sub-networks are smaller. Thus, restricting the growth space $\Lambda_k$ in each stage could help the convergence of the supernet. In practice, we make the restriction that $|\Lambda_k|\leq 9$. Specifically,
in the first stage, we use the largest, the smallest and the medium candidates of $n$ and $l$ in $\Omega$ to construct $\Lambda_1$, which makes it possible to route to the whole network and perform regular training if the growing ``ticket'' (suitable sub-network) does not exist. In each of the following stages, we include the next 3 candidates of $l$ and the next 1 candidate of $n$, forming a growth space with $2\times 4 = 8$ candidates.

\section{Additional Results}
We conduct additional experiments to explore our Advanced AutoProg framework, focusing primarily on AutoProg-\textit{One}. Some of the conclusions also apply to AutoProg-\textit{Zero}, such as the orthogonal speed-up achieved when combined with other acceleration methods.

\noindent\textbf{Theoretical Speedup.} In \cref{tab:appimagenet}, we calculate the average FLOPs per step of different learning schemes. AutoProg-\textit{One} consistently achieves more than 60\% speedup on theoretical computation. Remarkably, VOLO-D1 trained for 100 epochs with AutoProg-\textit{One} 0.4$\Omega$ achieves \textbf{132.2\%} theoretical acceleration. The gap between theoretical and practical speedup indicates large potential of AutoProg-\textit{One}. We leave the further improvement of practical speedup to future works; for example, AutoProg-\textit{One} can be further accelerated by adjusting the batch size to fill up the GPU memory during progressive learning.

\noindent\textbf{Comparison with Progressively Stacking.}
Progressively Stacking~\cite{Gong2019EfficientTO} (ProgStack) is a popular progressive learning method in NLP to accelerate BERT pretraining. It begins from $\frac{1}{4}$ of original layers, then copies and stacks the layers twice during training. Originally, it has three training stages with number of steps following a ratio 5:7:28. In CompoundGrow~\cite{Gu2021OnTT}, this baseline is implemented as three stages with 3:4:3 step ratio. Our implementation follows closer to the original paper, using a ratio of 1:2:5.
The results are shown in \cref{tab:progstack}. ProgStack achieves relatively small speedup with performance drop (0.4\%). Our MoGrow reduces this performance gap to 0.1\%. AutoProg-\textit{One} achieves 74.1\% more speedup and 0.5\% accuracy improvement over the ProgStack baseline.

\begin{table}[t]
    \centering
    \footnotesize
    \setlength{\tabcolsep}{4pt}
    \begin{tabular}{l|cc|c}
    \toprule
    Training scheme  & \makecell{Runtime\\(GPU hours)}& Speedup & Top-1 (\%)\\
    \midrule
    Baseline     & ~150.2~           & -& 82.6 \\
    \arrayrulecolor{lightgray}\hline\arrayrulecolor{black}
    ProgStack~\cite{Gong2019EfficientTO} &135.3 & +11.0\% & 82.2 \\
     + MoGrow & 136.0 & +10.4\% & 82.5 \\
     \arrayrulecolor{lightgray}\hline\arrayrulecolor{black}
     Prog & 93.3 & +60.9\% & 81.7\\
     \rowcolor{Light}AutoProg-\textit{One} 0.4$\Omega$  & 81.1     & \textbf{+85.1\%}& \textbf{82.7}\\
     \bottomrule
    \end{tabular}
    \caption{Comparison with progressively stacking.}%
    \label{tab:progstack}
\end{table}

\noindent\textbf{Adaptive Regularization.}
Adaptive Regularization (AdaReg) for progressive learning is proposed in \cite{Tan2021EfficientNetV2SM}. It adaptively change regularization intensity (including RandAug~\cite{Cubuk2020RandaugmentPA}, Mixup~\cite{Zhang2018mixupBE} and Dropout~\cite{Srivastava2014DropoutAS}) according to network capacity of CNNs. Here, we generalize this scheme to ViTs and study its effect on ViT AutoProg-\textit{One} training with DeiT-S and VOLO-D1. We mainly focus on three data augmentation and regularization techniques that are commonly used by ViTs, \ie, RandAug~\cite{Cubuk2020RandaugmentPA}, stochastic depth~\cite{Huang2016DeepNW} and random erase~\cite{Zhong2020RandomED}. When using AdaReg scheme, we linearly increase the magnitude of RandAug from 0.5$\times$ to 1$\times$ of its original value, and also linearly increase the probabilities of stochastic depth and random erase from 0 to their original values. The results of AutoProg-\textit{One} with and without AdaReg are shown in \cref{tab:ablation_AdaReg}. Notably, DeiT-S can not converge when training with AdaReg, probably because DeiT models are heavily dependent on strong augmentations. \textit{On the contrary}, AdaReg on VOLO-D1 is \textit{indispensable}. Not using AdaReg causes 1.2\% accuracy drop on VOLO-D1. This result is consistent with previous discoveries on CNNs~\cite{Tan2021EfficientNetV2SM}. By default, we use AdaReg on VOLO models and not use it on DeiT models.
\begin{table}[t]
\footnotesize
    \centering
    \begin{tabular}{l|c|c|c}
    \toprule
    Method              & AdaReg    & Speedup   & Top-1 Acc. (\%)  \\
    \midrule
    \rowcolor{Light}DeiT-S AutoProg-\textit{One}     & \xmark    &  \textbf{+40.7\%}  & \textbf{74.4}   \\
    DeiT-S AutoProg-\textit{One}     & \cmark    &   \na     & \pzo0.1$^*$    \\
    \midrule
    VOLO-D1 AutoProg-\textit{One}    & \xmark    &  +50.9\%  & 81.5  \\
    \rowcolor{Light}VOLO-D1 AutoProg-\textit{One}    & \cmark    &  \textbf{+85.1\%}  & \textbf{82.7}  \\

    \bottomrule
    \end{tabular}
    \caption{Ablation analysis of the adaptive regularization on ViTs with AutoProg-\textit{One}. (*: training can not converge)}
    \label{tab:ablation_AdaReg}
\end{table}

\noindent\textbf{Combine with AMP.}
Automatic mixed precision (AMP) [\textcolor{green}{52}] is a successful and mature low-bit precision efficient training method.
We conduct experiments to prove that the speed-up achieved by AutoProg-\textit{One} is orthogonal to that of AMP. As shown in \cref{tab:amp}, the relative speed-up achieved by AutoProg-\textit{One} with or without AMP is comparable (+85.1\% \vs +87.5\%), proving the orthogonal speed-up.%

\begin{table}[ht]
    \centering
    \footnotesize
    \setlength{\tabcolsep}{4pt}
    \begin{tabular}{l|c|c}
    \toprule
    Method          & Speed-up  & Top-1 Acc. (\%)  \\
    \midrule
    Original (w/o AMP)& \na       & 82.6 \\
    AMP             & +74.0\%   & 82.6 \\
    AutoProg-\textit{One}        &\cellcolor{Light} \textbf{+87.5\%}   & \textbf{82.7} \\
    \midrule
    \multirow{2}{*}{AMP + AutoProg-\textit{One}}  & \textbf{+222.1\%}   & \multirow{2}{*}{\textbf{82.7}}\\
    &\cellcolor{Light}(\textbf{+85.1\%} over AMP)&\\
    \bottomrule
    \end{tabular}
    \caption{Speed-up of AutoProg-\textit{One} is orthogonal to AMP [\textcolor{green}{52}].}
    \label{tab:amp}
\end{table}

\noindent\textbf{Number of stages.}
We perform experiments to analyze the impact of the number of stages on AutoProg-\textit{One} with different initial scaling ratios (0.5 and 0.4). As shown in \cref{tab:stage}, AutoProg-\textit{One} is not very sensitive to stage number settings. Fewer than 4 yields more speed-up, but could damage the performance. In general, the default 4 stages setting performs the best. When scaling the stage number to 50, there are only supernet training phases (2 epochs per stage) during the whole 100 epochs training, causing severe performance degradation.

\begin{table}[ht]
    \centering
    \footnotesize
    \setlength{\tabcolsep}{3pt}
    \begin{tabular}{c|l|c|c|c|c|c}
    \toprule
    Ratio & Num. Stages        & Orig.  & 3        & \cellcolor{Light}4        & 5             & 50  \\
    \midrule
    \multirow{2}{*}{0.5}&Speed-up        & \na    & \textbf{+69.1\%}   & \cellcolor{Light}+65.6\%  & +63.6\%       & +48.5\% \\
    &Top-1 Acc. (\%) & 82.6   & 82.6     & \cellcolor{Light}\textbf{82.8}     & \textbf{82.8}          & 81.7 \\
    \midrule
    \multirow{2}{*}{0.4}&Speed-up        & \na    & \textcolor{gray}{+90.8\%}   & \cellcolor{Light}\textbf{+85.1\%}  & +80.4\%       & \na \\
    &Top-1 Acc. (\%) & 82.6   & 82.4     & \cellcolor{Light}\textbf{82.7}     & \textbf{82.7}          & \na \\
    \bottomrule
    \end{tabular}
    \caption{Ablation analysis on number of stages.}
    \label{tab:stage}
\end{table}

\noindent\textbf{Effect of Progressive Learning in AutoProg-\textit{One}.} AutoProg-\textit{One} is comprised by its two main components, ``Auto'' and ``Prog''. The effectiveness of ``Auto'' is already studied by comparing with Prog in the main text. Here, we study the effectiveness of progressive learning in AutoProg-\textit{One} by training an elastic supernet baseline for 100 epochs without progressive growing to compare with AutoProg-\textit{One}. Specifically, we treat VOLO-D1 as an Elastic Supernet, and train it by randomly sampling one of its sub-networks in each step, same to the search stage in AutoProg-\textit{One}. The results are shown in \cref{tab:ablation_prog}. In previous works that uses elastic supernet~\cite{yu2019slimmable,Yu2020BigNASSU,chen2021autoformer}, the supernet usually requires more training iterations to reach a comparable performance to a single model. As expected, the supernet performance is lower than the original network given the same training epochs. Specifically, AutoProg-\textit{One} improves over elastic supernet baseline by 1.1\% Top-1 accuracy, with 17.1\% higher training speedup, reaching the performance of the original model with the same training epochs but much faster, which proves the superiority of progressive learning. 
\begin{table}[ht]
\setlength{\tabcolsep}{15pt}
\footnotesize
    \centering
    \begin{tabular}{l|c|c}
    \toprule
    Method      & Speedup           & Top-1 Acc. (\%)  \\
    \midrule
    Original    & \na               &   82.6    \\
    Supernet    & 48.5\%           &   81.7    \\
    \rowcolor{Light}AutoProg-\textit{One}    &\textbf{65.6\%}    & \textbf{82.8}\\
    \bottomrule
    \end{tabular}
    \caption{Ablation analysis of progressive learning in AutoProg-\textit{One} with VOLO-D1.}
    \label{tab:ablation_prog}
\end{table}

\noindent\textbf{Analyse of Searched Growth Schedule.}
Two typical growth schedules searched by AutoProg-\textit{One} are shown in \cref{tab:searched_schedule}. AutoProg-\textit{One} clearly prefers smaller token number than smaller layer number. Nevertheless, selecting a small layer number in the first stage is still a good choice, as both of the two schemes use reduced layers in the first stage.

\begin{table}[ht]
    \centering
    \footnotesize
    \begin{tabular}{l|c|c|c|c|c}
        \toprule
         \multicolumn{2}{c|}{Stage $k$}  & 1     &2      &3      &4  \\
         \midrule
         \multirow{2}{*}{\makecell{VOLO-D1 100e 0.4$\Omega$}}&$l$        & 0.4   &1      &1      &1  \\
         \arrayrulecolor{lightgray}\cline{2-6}\arrayrulecolor{black}
         &$n$        & 0.4   &0.6    &0.6    &1  \\
        \hline
        \multirow{2}{*}{\makecell{VOLO-D2 300e}}&$l$        & 0.83  &1      &1      &1  \\
        \arrayrulecolor{lightgray}\cline{2-6}\arrayrulecolor{black}
         &$n$        & 0.5   &0.67   &0.83   &1  \\
         \bottomrule
    \end{tabular}
    \caption{Searched growth schedules for VOLO-D1 0.4$\Omega$, 100 epochs, and VOLO-D2, 300 epochs.}
    \label{tab:searched_schedule}
\end{table}

\noindent\textbf{Retraining with Searched Growth Schedule.}
To evaluate the searched growth schedule, we perform retraining from scratch with VOLO-D1, using the schedule searched by AutoProg-\textit{One} 0.4$\Omega$. As shown in \cref{tab:retrain}, retraining takes slightly longer time (-0.6\% speedup) because the speed of searched optimal sub-networks could be slightly slower than the average speed of sub-networks in the elastic supernet. Retraining reaches the same final accuracy, proving that the searched growth schedule can be used separately.

\begin{table}[ht]
    \centering
    \footnotesize
    \setlength{\tabcolsep}{4pt}
    \begin{tabular}{l|cc|c}
    \toprule
    Training scheme     &\makecell{Runtime\\(GPU hours)} & Speedup & Top-1 (\%) \\
    \midrule
        Baseline    & ~150.2~           & -& 82.6 \\
    \arrayrulecolor{lightgray}\hline\arrayrulecolor{black}
    \rowcolor{Light}AutoProg-\textit{One} 0.4$\Omega$ & 81.1     & \textbf{+85.1\%}& \textbf{82.7}\\
    Retrain & 81.4 & +84.5\% & \textbf{82.7}\\
    \bottomrule
    \end{tabular}
    \caption{Retraining results with searched growth schedule on VOLO-D1, 100 epochs.}
    \label{tab:retrain}
\end{table}

\noindent\textbf{Extend to CNNs.} To explore the effect of our policy on CNNs, we conduct experiments with ResNet50 \cite{he2016deep}, and found that the policy searched on ViTs generalizes very well on CNNs (see \cref{tab:r50}). These results imply that AutoProg-\textit{One} opens an interesting direction (automated progressive learning) to develop more general learning methods for a wide computer vision field.

\begin{table}[ht]
    \centering
    \footnotesize
    \setlength{\tabcolsep}{13pt}
    \begin{tabular}{l|c|c}
    \toprule
    Method          & Speed-up  & Top-1 Acc. (\%)  \\
    \midrule
    Original        & \na       & 77.3 \\
    \rowcolor{Light}AutoProg-\textit{One}        & \textbf{+56.9\%}   & 77.3 \\
    \bottomrule
    \end{tabular}
    \caption{AutoProg-\textit{One} with ResNet50 \cite{he2016deep} on ImageNet (100 epochs).}
    \label{tab:r50}
\end{table}

\end{document}